\documentclass[journal]{IEEEtran}
\usepackage{amsmath,amsfonts}
\usepackage{array}
\usepackage[caption=false,font=normalsize,labelfont=sf,textfont=sf]{subfig}
\usepackage{textcomp}
\usepackage{stfloats}
\usepackage{url}
\usepackage{multirow} 
\usepackage{verbatim}
\usepackage{cite}
\usepackage{tikz}
\usetikzlibrary{matrix}
\usepackage[ruled,linesnumbered]{algorithm2e}
\usepackage{lipsum}
\usepackage{epsfig}

\usepackage{graphicx}

\usepackage[noend]{algpseudocode}
\usepackage{comment}

\usepackage{booktabs, multirow} 
\usepackage{soul}

\usepackage{fontawesome}
\usepackage{lipsum}
\usepackage{verbatim}
\usepackage{notoccite}  
\usepackage{hyperref}\usepackage[caption=false,font=normalsize,labelfont=sf,textfont=sf]{subfig}
\usepackage{makecell}
\usepackage{threeparttable}
\title{Sim-Suction: Learning a Suction Grasp Policy for Cluttered Environments Using a Synthetic Benchmark   }

\author{Juncheng Li$^{1}$, David J. Cappelleri$^{1,2}$ 
\thanks{$^{1}$ J. Li and D. Cappelleri are with the Multi-Scale Robotics \& Automation Lab, School of Mechanical Engineering, Purdue University, West Lafayette, IN USA.  $^{2}$ D. Cappelleri is also with the Weldon School of Biomedical Engineering (By Courtesy), Purdue University, West Lafayette, IN USA.
{\tt\small \{li3670, dcappell\}@purdue.edu}}
}
\begin{document}

\markboth{IEEE TRANSACTIONS ON ROBOTICS}%
{IEEE TRANSACTIONS ON ROBOTICS}
\IEEEoverridecommandlockouts
\IEEEpubid{\makebox[\columnwidth]{ 
    1552-3098~\copyright~2023 IEEE 
    \hfill} \hspace{\columnsep}\makebox[\columnwidth]{ }}

\maketitle
\IEEEpubidadjcol

\begin{abstract}
This paper presents \textit{Sim-Suction}, a robust object-aware suction grasp policy for mobile manipulation platforms with dynamic camera viewpoints, designed to pick up unknown objects from cluttered environments. Suction grasp policies typically employ data-driven approaches, necessitating large-scale, accurately-annotated suction grasp datasets. However, the generation of suction grasp datasets in cluttered environments remains underexplored, leaving uncertainties about the relationship between the object of interest and its surroundings. To address this, we propose a benchmark synthetic dataset, \textit{Sim-Suction-Dataset}, comprising 500 cluttered environments with 3.2 million annotated suction grasp poses. The efficient \textit{Sim-Suction-Dataset} generation process provides novel insights by combining analytical models with dynamic physical simulations to create fast and accurate suction grasp pose annotations. We introduce \textit{Sim-Suction-Pointnet} to generate robust 6D suction grasp poses by learning point-wise affordances from the \textit{Sim-Suction-Dataset}, leveraging the synergy of zero-shot text-to-segmentation. Real-world experiments for picking up all objects demonstrate that \textit{Sim-Suction-Pointnet} achieves success rates of 96.76\%, 94.23\%, and 92.39\% on cluttered level 1 objects (prismatic shape), cluttered level 2 objects (more complex geometry), and cluttered mixed objects, respectively. The codebase can be accessed at \href{https://github.com/junchengli1/Sim-Suction-API}{\textit{https://github.com/junchengli1/Sim-Suction-API}}.

\end{abstract}

\section{INTRODUCTION}

\begin{figure}[ht]
\centering
\includegraphics[width=0.988\linewidth]{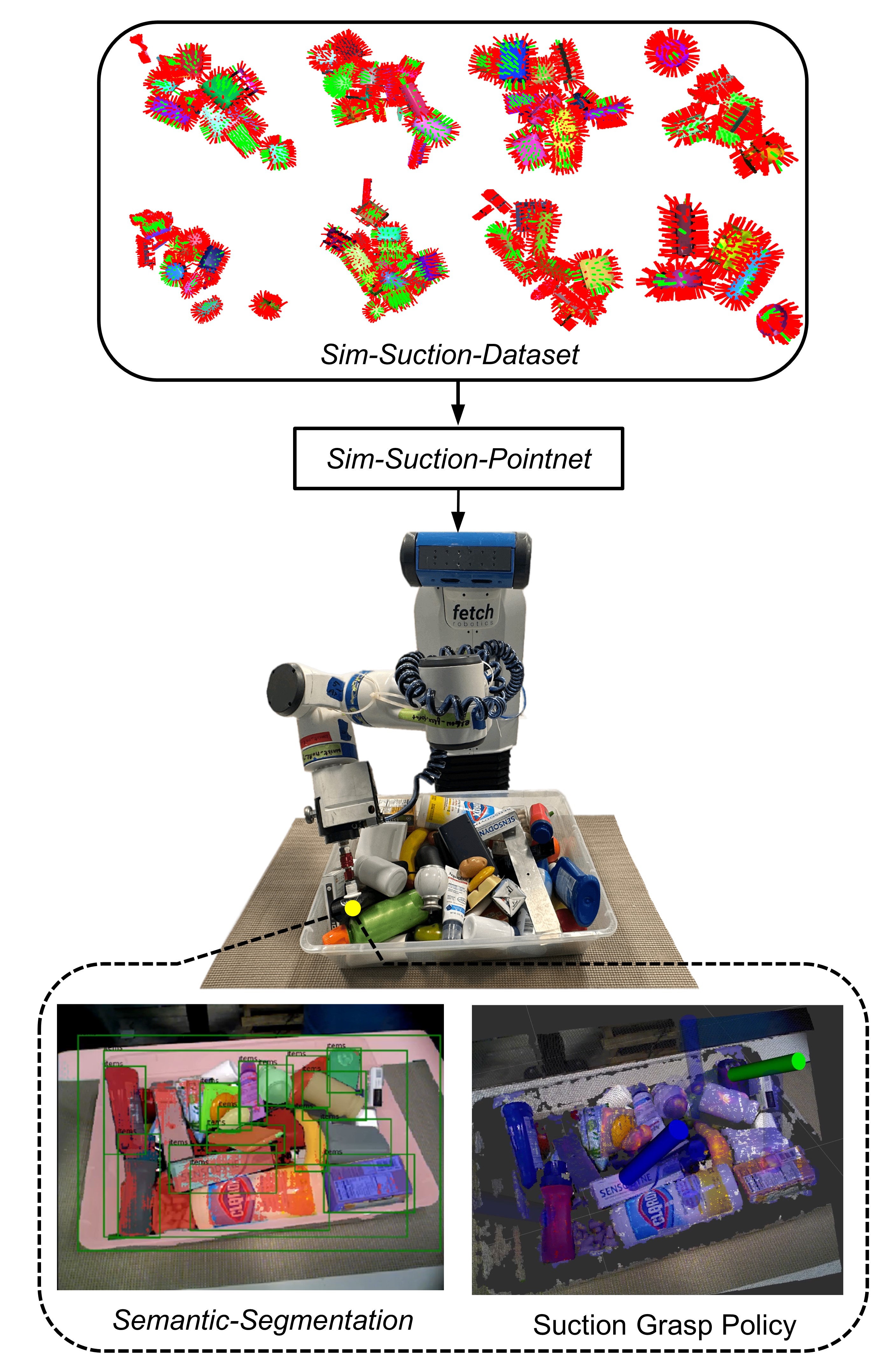}
\vspace{-0.10in}
\caption{\footnotesize Overview of \textit{Sim-Suction}. The \textit{Sim-Suction} is a deep-learning based policy to determine the robust suction grasp poses in cluttered environments. It has the following components: \textit{Sim-Suction-Dataset}, a large-scale synthetic dataset for suction cup gripper that combines analytical model and physical simulation; \textit{Sim-Suction-Pointnet}, an object-aware point-wise affordance network that uses text prompt to predict grasp success probability for given picking-up task.}
\label{fig:Sim-Suction}
\vspace{-0.1in}
\end{figure}

\IEEEPARstart{T}{he} development of autonomous mobile manipulation platforms is crucial for the future of space habitats, where robots can perform various tasks in cluttered environments with minimal human intervention. In these habitats, tasks such as maintenance, cargo handling, and assembly of structures have unique challenges for grasping and manipulation due to confined spaces, limited resources, and the need to handle objects with diverse shapes, sizes, and materials. Furthermore, space habitats often contain cluttered environments with objects that may be partially occluded or challenging to access. While space habitats represent a vital application area, mobile manipulation platforms also play an essential role in industry 4.0 and household settings due to their flexibility and efficiency. However, humans expect mobile manipulation platforms to be fully autonomous without any intervention. This is challenging for manipulation tasks, where robots have trivial or no pre-existing knowledge, unlike tasks on a predictable assembly line under controlled conditions. For example, typical dynamic manipulation tasks may include picking objects from a bin, cleaning a cluttered workbench, or retrieving products from a shelf. These tasks are challenging for mobile manipulation platforms, both in terms of identifying the grasp region and executing the mechanical grasping process. Unlike familiar objects, novel objects are items that the robot has never encountered before, hence, no prior information about their shape, size, weight, texture, and other physical properties is available. To solve these tasks, mobile manipulation platforms require the ability to observe the cluttered environments, decide on the way to grasp the object of interest, and perform a robust grasp once found. This is a difficult task due to the challenges associated with where to grasp and the uncertainties on how objects with varying size, weight, shape, and texture will react when trying to establish grasp contact point. The mobile manipulation platforms need to understand the task requirements that humans give and make the right decisions in such settings. Our work in this paper aims to tackle the picking-up challenges by using text prompts to guide the robot in completing tasks such as picking up all novel objects from a bin or selecting specific novel objects from the bin based on a brief text description.

Suction cup grippers play a vital role in warehouses due to their simplicity, compactness, light weight, and minimal maintenance requirements. They can also handle a wide range of objects, from fragile parts to large dimension objects. The Amazon Picking Challenge showed that the suction cup gripper is commonly used for general picking tasks with a higher success rate than other grippers~\cite{amazon_pick}. Experiments from DexNet 4.0~\cite{dexnet4.0} also demonstrate the preference for choosing a suction gripper over a parallel jaw gripper, with an 82\% selecting rate on the bin-picking task. Previous studies show that the suction cup gripper outperforms other grippers in successfully grasping objects from cluttered environments due to its ability to create a single contact point on the object surface through a narrow space. Suction grippers are particularly suited for handling objects in space habitats with various surface properties, as they can establish a secure grip on a wide range of materials without causing damage. This is particularly important for delicate equipment and components that require gentle handling and precise placement. Our previous work designed a modular end-effector system~\cite{modular_end_effector}, which enables a mobile manipulation platform to use a suction cup gripper more efficiently with an embedded vacuum generator and control module. In this work, we focus on developing the suction grasping policy for mobile manipulation platforms to tackle the challenge of grasping objects from a cluttered environment.

In the grasping research community, a large number of studies pay particular attention to the parallel-jaw gripper grasping policy. At the same time, a relatively small body of literature is concerned with the suction grasping policy. Studies on suction grasping\cite{dexnet3,zeng,suctionnet,jiang2023multipleobject} have begun to examine the data-driven approaches\cite{datadriven1,datadriven2,datadriven3,datadriven4} using deep learning and achieve better performances over traditional online heuristic baseline approaches. The data-driven approaches can be classified into training on realistic datasets \cite{zeng},\cite{suctionnet} and training on synthetic datasets~\cite{dexnet3}. The key problem of realistic datasets collected from human or real experiments\cite{real_experiment} is that the time cost to retrieve object information is large\cite{cornell}. It is also difficult to get object instance masks and their 6D pose information due to occlusion. Therefore realistic datasets are usually small in size and have sparse information. The human labeling process \cite{zeng} is another potential concern because it is hard to generalize to other systems, and the accuracy of annotating suction grasp poses as ground truth is questionable. More recent training on synthetic dataset approaches can help reduce the cost of data collection but still have limitations regarding the analytical model accuracy, domain gap, fixed vision system, neglecting objects and suction cup gripper dynamics, and insufficient information on cluttered environments. Furthermore, different neural network-based learning methods ~\cite{dexnet3,zeng,suctionnet,jiang,zhang} for suction cup grippers have been proposed to predict the grasp success using pixel-wise affordance with RGB or depth inputs. These 2D affordance methods rely on images that experience some issues with generalization to unseen objects. Learning on 3D point clouds provides better generalization~\cite{pointnetgpd} for novel objects. However, to our knowledge, no study on suction cup grasp success prediction uses object-aware point-wise affordance, which directly takes the 3D point cloud and text prompt as input and generates robust 6D suction grasp poses for object instances. Additionally, the current evaluation metrics \cite{dexnet4.0},\cite{zeng} for suction grasp prediction precision by comparing the pre-annotated suction grasp ground truth with the prediction result, suffer from the fact that the pre-annotated suction grasp dataset can only include a subset of all possible candidates depending on the sampling number, due to the nature of infinite suction grasp poses existing in every cluttered environments\cite{billionway}. Together, these limitations demand a universal, accurate, and efficient dataset generation process and network architecture and evaluation metrics, which can serve as a benchmark for developing suction grasp policies.   

In this paper, we propose \textbf{\textit{Sim-Suction}}, a deep learning-based system that uses a suction cup gripper to pick up novel objects from cluttered environments (Fig.~\ref{fig:Sim-Suction}). It consists of two components: (1) \textbf{\textit{Sim-Suction-Dataset}}: a large-scale synthetic dataset for cluttered environments, and (2) \textbf{\textit{Sim-Suction-Pointnet}}: an object-aware point-wise affordance policy that predicts the most robust suction grasp pose for the target object. The primary contributions of our work include:
{
\begin{itemize}
    \item A large-scale synthetic dataset for cluttered environments that include RGB images, depth information, single-viewed point clouds, multi-viewed point clouds, object instance segmentation masks, 6D object poses, 2D object bounding boxes, 6D object bounding boxes, camera matrices, 6D suction grasp poses, and 3D suction grasp score maps.   
    \item A suction grasp candidate evaluation process that integrates an analytical model and simulation, assessing object collision, seal formation, suction cup gripper dynamics, and multi-body rigid dynamics in cluttered environments.
    \item Physical robot experiments validating the Sim-Suction analytical suction model and comparing it with the DexNet model.
    \item A novel point-wise affordance network that trains on point clouds and annotated 3D suction grasp score map, which outputs point-wise suction grasp success probability.
    \item An online evaluation metric capable of assessing the precision of suction grasp predictions across different benchmarks. 
    \item A thorough ablation study examining the effectiveness of the \textit{Sim-Suction-Dataset} diversity and \textit{Sim-Suction-Pointnet} architecture.
    \item Simulation and physical robot experiments quantifying the \textit{Sim-Suction-Pointnet} suction grasp success rate without prior knowledge of objects in cluttered environments.
\end{itemize}
}

\section{RELATED WORK}
\subsection{Object Affordance}
Object affordance was first introduced by Gibson \textit{et al.}\cite{affordance}, which refers to the ability of an agent to perform actions in a given environment. In the suction grasping community, the vast majority of studies use object suction affordance~\cite{dexnet3},~\cite{zeng},~\cite{suctionnet},~\cite{jiang},~\cite{zhang} to indicate the most likely part of objects to make the suction grasp successful based on the unique mechanism and shape of a suction cup. Affordance learning is a variant of the segmentation method, which learns from the suction affordance scores and can adapt to novel objects. Research on the suction affordance learning framework mainly focuses on the FCN-based pixel-wise affordance\cite{fcn} that only uses RGB-D images or depth to infer the suction grasp success probability at each pixel. UMPNet~\cite{UMPNet} proposes an image-based policy network that infers closed-loop action sequences for manipulating articulated objects. However, image-based affordance often faces generalization challenges when encountering unseen images. Recent studies have started to explore one-shot~\cite{oneshot,oneshot2} or zero-shot~\cite{zeroshot1} image-based affordance to address generalization issues in object grasping, but these methods necessitate extensive datasets and resources for training. With the development of PointNet~\cite{pointnet} and PointNet++~\cite{pointnet++}, the feature extractor can extract the 3D features directly from raw point cloud inputs that have good performance in unseen objects. A number of studies~\cite{pointnetgpd},\cite{6D-grasp},\cite{pointnetgrasp},\cite{graspnet},\cite{o2o},\cite{catgrasp} begin to learn point-wise graspable affordance for parallel jaw grippers, but no research has been found that investigate point-wise suction affordance. 


\subsection{Semantic Segmentation}

Semantic segmentation is an essential component in various robotic applications, particularly object grasping. Several studies have used segmentation masks to facilitate grasp planning~\cite{segment1,segment2,segment3,segment4,segment5,dexnet4.0}. By providing a detailed representation of object boundaries and spatial relationships, semantic segmentation enables robots to understand the shape, size, and category of objects. This information is essential for calculating feasible grasp points and optimizing grasp strategies. Mask R-CNN~\cite{maskrcnn} has been widely used in robotics applications, such as object grasping, due to its ability to provide precise object localization and segmentation. However, Mask R-CNN does not generalize to unseen objects in novel categories. Recent ground-breaking work promptable Segment Anything Model (SAM)~\cite{SAM} demonstrates promising out-of-box zero-shot image segmentation capabilities in various scenarios without any retraining and fine-tuning. It requires 2D points or 2D bounding boxes prompts to provide instance segmentation. The Grounding DINO model~\cite{groundingdino} proposes open-set image object detector which incorporates a language model to enhance concept understanding, resulting in more effective object detection for unseen objects. Our work fuses the advantages of SAM~\cite{SAM}, Grounding DINO~\cite{groundingdino}, and point-wise affordance to propose a suction-grasping policy to use text prompt input to guide the picking-up task by predicting robust 6D suction grasp poses for objects of interest in cluttered environments.

\subsection{Suction Grasp Dataset}
Zeng \textit{et al.}~\cite{zeng} proposed a manually labeled suction grasp dataset from cluttered real-world environments. It requires humans with experience to annotate each pixel in the RGB-D images with a binary value representing suctionable and non-suctionable areas. One major drawback of this dataset is that the dataset size is relatively small and include limited object information. Moreover, the empirical suctionable area labeling process is tedious and introduces potential errors. SuctionNet-1billion~\cite{suctionnet} addresses this issue by proposing a real-world suction grasp dataset and using the analytical model to annotate RGB-D images captured from two popular cameras. However, the time cost to generate a rich real-world dataset remains a considerable limitation. It is also hard to generalize to different environments and vision settings. Dexnet 3.0~\cite{dexnet3} instead generates a synthetic suction grasp dataset and uses an analytical model to annotate on singulated object depth images, which do not contain any information about the cluttered environments. Jiang \textit{et al.}~\cite{jiang} turns to generate a synthetic dataset in cluttered environments. However, it fails to consider the domain gap and provides a limited annotation method by analyzing primitive shapes only. Shao \textit{et al.}~\cite{Shao2019SuctionGR} proposes a suction grasp dataset used for self-supervised learning in cluttered environments but only contains cylinders of the same size, which is restricted to specific applications. 

\subsection{Grasp Candidates Sampling}
The grasp candidates sampling process refers to randomly sampling the configurations of the end-effector on the target object to generate a large number of possible grasp configurations, for which the grasp 6D grasp poses cannot be calculated directly. Most research in suction grasp sampling has been carried out to sample suction grasp candidates in point cloud space \cite{dexnet3},\cite{suctionnet}, with a little focus on sampling suction grasp candidates on 2D images\cite{object_agnostic_suction}. Our main focus is to look at grasp sampling in point cloud space since 6D suction grasp poses are highly dependent on the geometry of the objects. The grasp candidates sampling process in point cloud space can be classified into object-agnostic sampling and object-aware sampling. The object-agnostic sampling process does not need individual object information. It treats multiple objects in a cluttered environment as a single unified object. Object-agnostic sampling algorithms\cite{object_agnostic_suction}  perform a search on the entire point cloud space, which is slow and inaccurate in cluttered environments. The object-aware sampling process can solve the above issues by using the complete information in the entire point cloud and help with further evaluation. In this work, we use our dataset's instance segmentation mask and object 6D poses to create an object-aware sampling scheme, where each suction grasp candidate is associated with the relevant object information.

\subsection{Suction Grasp Candidates Evaluation}

For evaluating suction cup grasp candidates, it is challenging to annotate good and bad suction grasps. Mahler\textit{et al.}~\cite{dexnet3} first propose a compliant suction contact model which uses a spring system to evaluate the seal formation on the contact surface and quasi-static physics to evaluate the ability to resist external wrenches for the singulated object. Cao \textit{et al.}~\cite{suctionnet} extended the work by simplifying the quasi-static spring system for checking seal formation and resisting external wrenches on the sigulated object, and performed collision checking in cluttered environments. Zhang \textit{et al.}~\cite{zhang} adopts a similar model to evaluate suction grasp candidates on singulated object but fails to address constraints in cluttered environments. Jiang \textit{et al.}~\cite{jiang} uses a convolution-based method to calculate the suctionable area, assuming the suctionable surface is flat and large enough. The authors evaluate the ability to resist external wrenches by only calculating the normalized distance between the suction location and the center of the suctionable area of each object. Overall, the current suction quasi-static physics model has limitations in cluttered environments because it cannot comprehensively analyze the entire cluttered environment and that can lead to false results, especially when the suction cup gripper tries to grab objects from the bottom of the heap when other objects are stacked on them. It also fails to analyze whether the contact is kept established during the suction cup gripper movement. The current seal formation evaluations only examine the contact surface of a singulated object. However, in cluttered environments, a suction cup may have contact with multiple objects or the ground plane when the suction location is on object edge. The studies presented thus far demand the need for an accurate suction grasp candidates evaluation scheme in cluttered environments. This paper makes an essential contribution to suction grasping in cluttered environments by focusing on the entire suction grasp process, including suction cup gripper and object dynamics, instead of only analyzing the quality of established contacts for singulated object. We combine the analytical model with physics simulations to provide an accurate suction grasp evaluation process, which also serves as an online evaluation metric to calculate the prediction accuracy.

\section{PROBLEM STATEMENT}

\subsection{Overview}
We denote the unstructured environment initial state as $\mathcal{X}=(\mathcal{O},\mathcal{P},\mathcal{C},\mathrm{d})$. Given a single view RGB-D image and registered point cloud $\mathcal{P}$ of an unstructured environment consisting of a set of objects $\mathcal{O}$ captured from a depth camera with known camera matrix $\mathcal{C}$, our goal is to enable the robot to use a vacuum suction cup gripper with suction pad diameter $\mathrm{d}$ to pick up object $\mathcal{O}_i$ from $\mathcal{O}$ by selecting the most robust suction cup grasp pose
$\mathrm{S}$=($\mathrm{R}$,$\mathrm{T}$) $\in$ SE(3) from all possible suction pose candidates $\mathbb{S}$, where $\mathrm{R}$ represents the suction cup approaching direction and $\mathrm{T}$ represents the location of the center of the suction pad.

The binary success measurement of picking up $\mathcal{O}_i$ from $\mathcal{O}$ depends on the state of $\mathcal{O}_i$, each object state specifies the object geometry, the center of mass, 6-DoF pose, friction coefficient, and interactions with its surroundings. Therefore, we need to consider those constraints to predict a successful suction grasp. Data-driven approaches showed their capability to handle these physics constraints by either using human labeling or analytical models. However, limitations still exist regarding the data size, the precision of the analytical model, the ability to generalize to customized objects, and the performance in cluttered contexts. To address the aforementioned limitations, we first propose a method to autonomously create and annotate a large-scale synthetic dataset for cluttered environments using the Omniverse Isaac Sim simulator~\cite{isaac}. This results in the creation of a benchmark dataset called \textit{Sim-Suction-Dataset}. Subsequently, we train the dataset with affordance inference networks, \textit{Sim-Suction-Pointnet}, to generate a point-wise suction grasp affordance map. This map is then combined with a task-oriented semantic segmentation mask to generate the grasping policy and refine 6D suction poses for target objects.

\subsection{Assumptions}
We make the following assumptions when developing the \textit{Sim-Suction}:
\begin{itemize}
  \item Objects $\mathcal{O}$ are rigid bodies made of non-porous material with mass, inertia, velocity, and friction and can interface with static or moving rigid bodies in the unstructured environment;
  \item The suction gripper can be simulated with a spring-mass model and can create a 6-DoF joint with the object of interest. 
  \item The depth camera has known intrinsic matrix $\mathcal{C}$.
 \end{itemize}

\section{Sim-Suction-Dataset}
\begin{figure}
\centering
\includegraphics[width=1\linewidth]{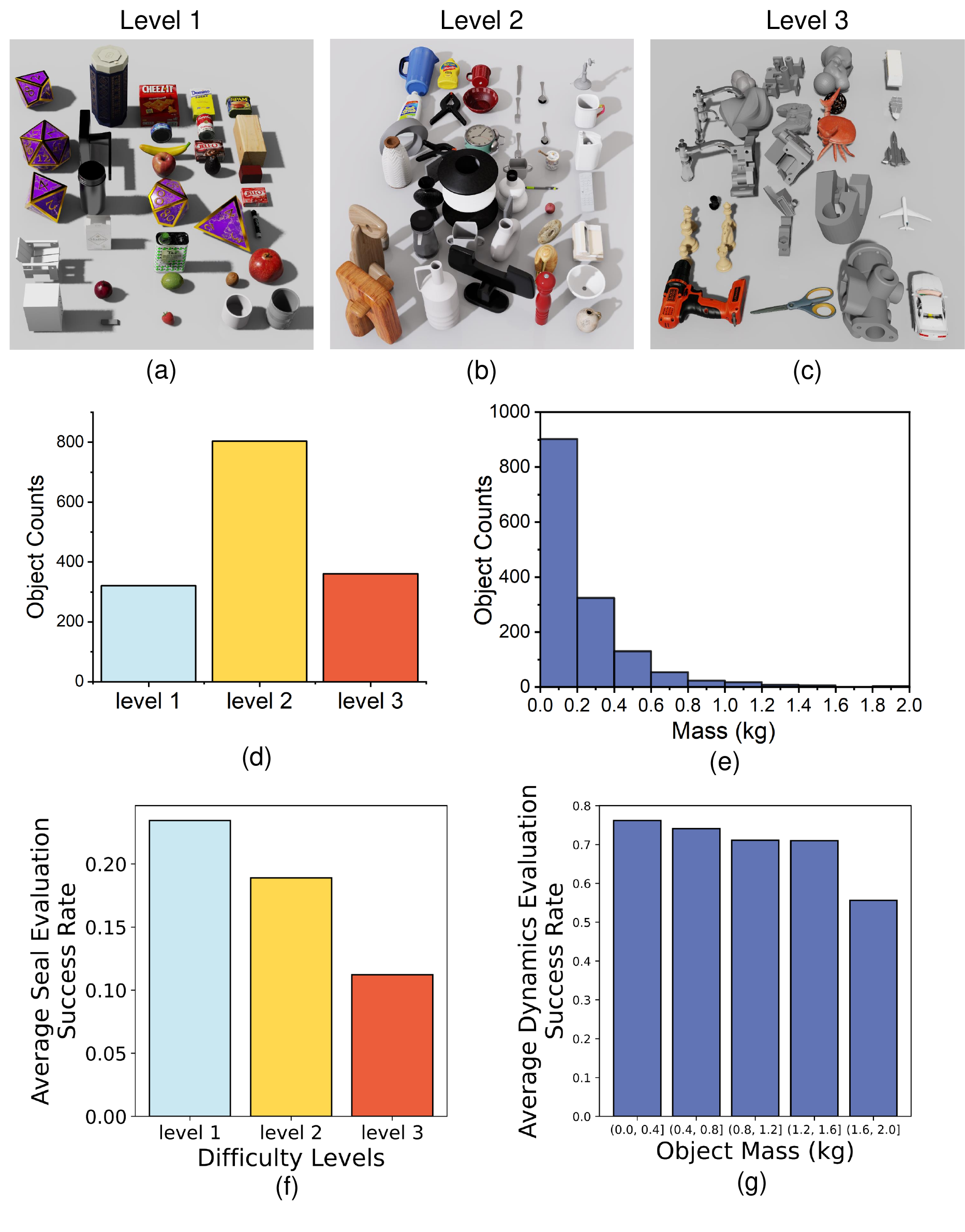}
\vspace{-0.20in}
\caption{\footnotesize (a) - (c) Examples of objects with varying difficulty levels, where Level 1 is the least challenging and Level 3 is the most challenging. (d) - (e) Depiction of object mass and difficulty levels distribution within the Sim-Suction-Dataset. (f) Influence of object difficulty levels on seal evaluation. (g) Effect of object mass on dynamics evaluation.}
\label{dataset_study}

\end{figure}

\begin{figure}
\centering
\includegraphics[width=1\linewidth]{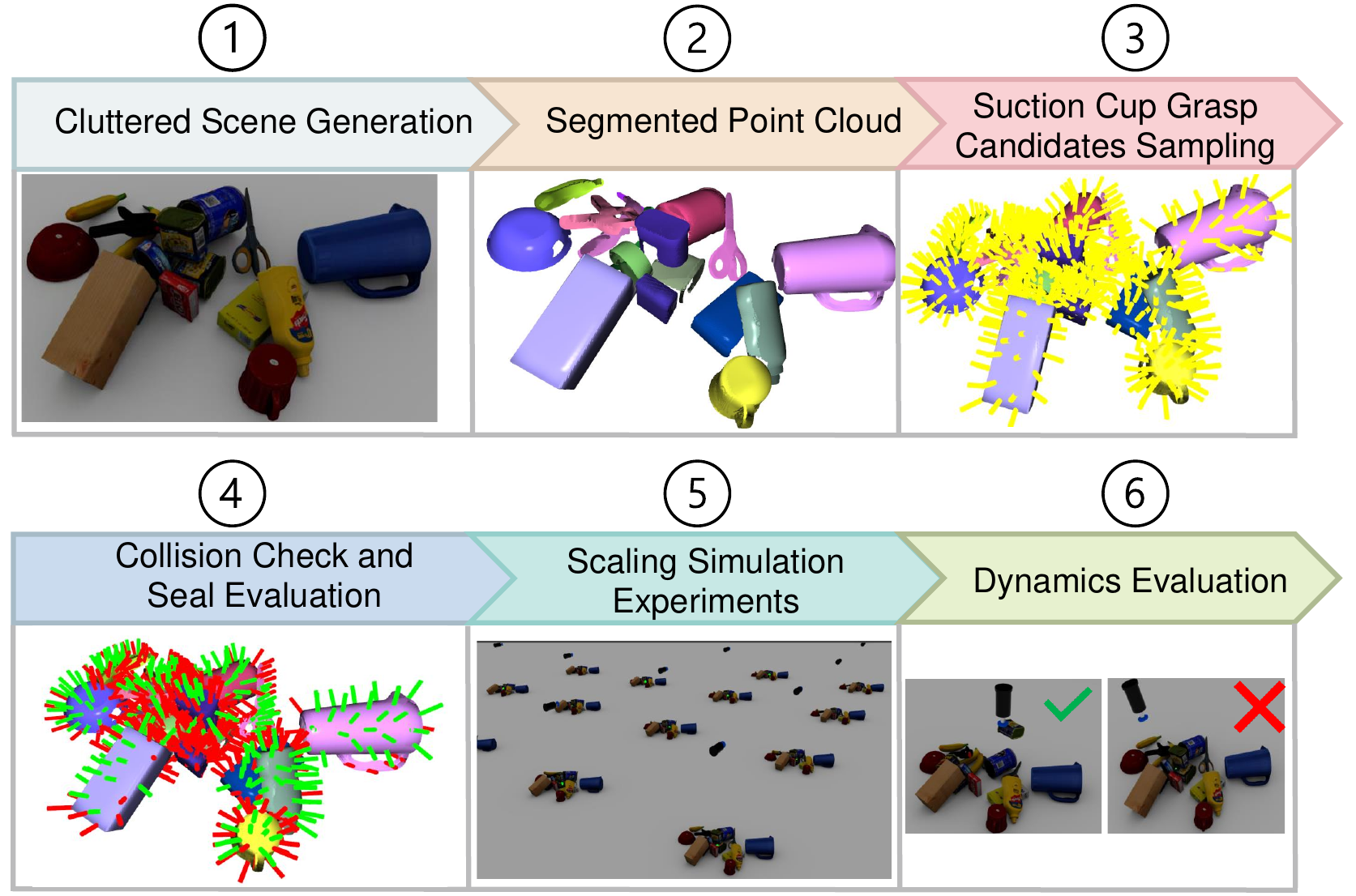}
\vspace{-0.10in}
    \caption{\footnotesize Sim-Suction-Dataset generation pipeline. \raisebox{.5pt}{\textcircled{\raisebox{-.9pt} {1}}} Objects free fall above the ground plane to create a cluttered scene. \raisebox{.5pt}{\textcircled{\raisebox{-.9pt} {2}}} Create segmented point clouds for each scene from multi-viewed synthetic camera. \raisebox{.5pt}{\textcircled{\raisebox{-.9pt} {3}}} Sample suction grasp candidates from the object surface. \raisebox{.5pt}{\textcircled{\raisebox{-.9pt} {4}}} Evaluate each candidate with a combination of the analytical model and PhysX engine to check whether it fails to form a seal or has collisions with surroundings. \raisebox{.45pt}{\textcircled{\raisebox{-.7pt} {5}}}-\raisebox{.45pt}{\textcircled{\raisebox{-.7pt} {6}}} Use simulation to further evaluate each candidate under realistic dynamics settings.}
\label{pipeline}

\end{figure}

    Previous suction grasp datasets mainly focused on the labor-intensive collection and labeling processes, which limit dataset size and ground truth precision. Although analytical models like Dexnet-3.0 can reduce some errors introduced by human labeling, they can only be applied to singulated objects and fail to consider dynamic interactions in unstructured environments. Additionally, real-world datasets do not transfer well to different scenarios due to fixed lighting conditions, camera systems, and difficulties in adding new objects. 
    To overcome these issues, we present methods for generating a large-scale dataset through physics simulation, resulting in the \textit{Sim-Suction-Dataset}. This dataset is the first large-scale synthetic suction grasp dataset in cluttered environments that combines analytic models and dynamic interactions. Our dataset (see Table.~\ref{table:Comparison of the suction grasp dataset}) includes 1550 objects from 137 categories. The objects come from ShapeNet~\cite{shapenet}, YCB objects~\cite{ycb}, NVIDIA Omniverse Assets~\cite{nvidia}, and Adversarial Objects in Dex-Net~\cite{dexnet1}. The objects can be categorized into three difficulty levels (Fig.~\ref{dataset_study}): Level 1 includes prismatic and circular solids, Level 2 includes objects with varied geometry, and Level 3 includes objects with adversarial geometry and material properties. The resulting mass and difficulty level distribution are shown in Fig.~\ref{dataset_study}. (d) and (e). The difficulty levels of the objects are determined based on the complexity of their triangle mesh. We use the material density of each object to calculate the object's mass. We select a 1.5 cm radius bellows suction cup and a 2.5 cm radius bellows suction cup for our suction cup gripper. We propose a pipeline to study the grasp correlation of each object in the entire cluttered environment to automate the generation of accurately labeled data. This is achieved by combining sampling-based approaches, analytical model analysis, domain randomization, and dynamic physics simulations. We use Nvidia Omniverse Isaac Sim Simulator~\cite{isaac} with the built-in PhysX engine~\cite{tracing} as a toolkit to simulate rigid body dynamics and annotate 6D suction grasp poses(Fig.~\ref{pipeline}). Each grasp has a corresponding 6D grasp pose, gripper dimension, object difficulty level, object physics information, RGB-D image, camera matrix, binary success label, and object segmentation point cloud associated with it. This comprehensive dataset and auto-labeling pipeline are intended to serve as a synthetic benchmark and reference for suction grasping research. Notably, the dataset generation pipeline itself could serve as a foundational model-based strategy. This opens doors for researchers to sample suction grasps without a dedicated trained model. Our dataset provides a common dataset and approach for comparison and evaluation of suction grasping algorithms across different gripper sizes and can also be used to develop point-wise affordance grasping policies (Section~\ref{sec:pointnet}).
    
\begin{table*}[h]
\caption{Comparison of Suction Grasping Datasets.}\label{table:Comparison of the suction grasp dataset}
\vspace{-0.10in}
\scalebox{1}{
\begin{tabular}{@{}cccccccccc@{}}
\toprule
\multirow{2}{*}{\textbf{Dataset}} & \textbf{Grasp Pose}     & \textbf{Objects/} & \textbf{Camera} & \textbf{Total}   & \textbf{Total}  &\multirow{2}{*}{\textbf{Modality}}       & \textbf{Multiple Gripper} & \textbf{Semantic}     & \textbf{Dynamics}   \\
                                  & \textbf{Label (Method)} & \textbf{Scene}    & \textbf{Type}   & \textbf{Objects} & \textbf{Labels} &      & \textbf{sizes}            & \textbf{Segmentation} & \textbf{Evaluation} \\ \midrule
               
SuctionNet~\cite{suctionnet}                       & 6D (\faPencil )                  & $\sim$10          & Real            & 88               & $\sim$1.1B      & RGB-D                 & No                        & Yes                   & No                  \\
Dex-Net 3.0~\cite{dexnet3}                       & 2D (\faPencil )                  & 1              & Sim             & 1.6K             & $2.8$M            & Depth          & No                        & No                    & No                  \\
A. Zeng~\cite{zeng}                            & 2D (\faUser )                  & NA                & Real            & NA               & 191M            & RGB-D          & No                        & No                    & No                  \\
\textit{\textbf{Sim-Suction-Dataset}}        & \textbf{6D (\faPencil , \faPlay )}      & \textbf{1-20}     & \textbf{Sim}    & \textbf{1.5K}    & \textbf{~3.2M}   & \textbf{RGB-D} & \textbf{Yes}              & \textbf{Yes}          & \textbf{Yes}        \\ \bottomrule
\end{tabular}
}
\begin{tablenotes}
      \item Note: Grasp labels can be generated either manually (\faUser), using analytical models (\faPencil), or through physics simulation (\faPlay). Dynamics evaluation is denoted as partial when it only evaluates an isolated single object, rather than objects in cluttered environments.
    \end{tablenotes}
    \vspace{-0.20in}
\end{table*}

\begin{figure}
\centering
\includegraphics[width=1\linewidth]{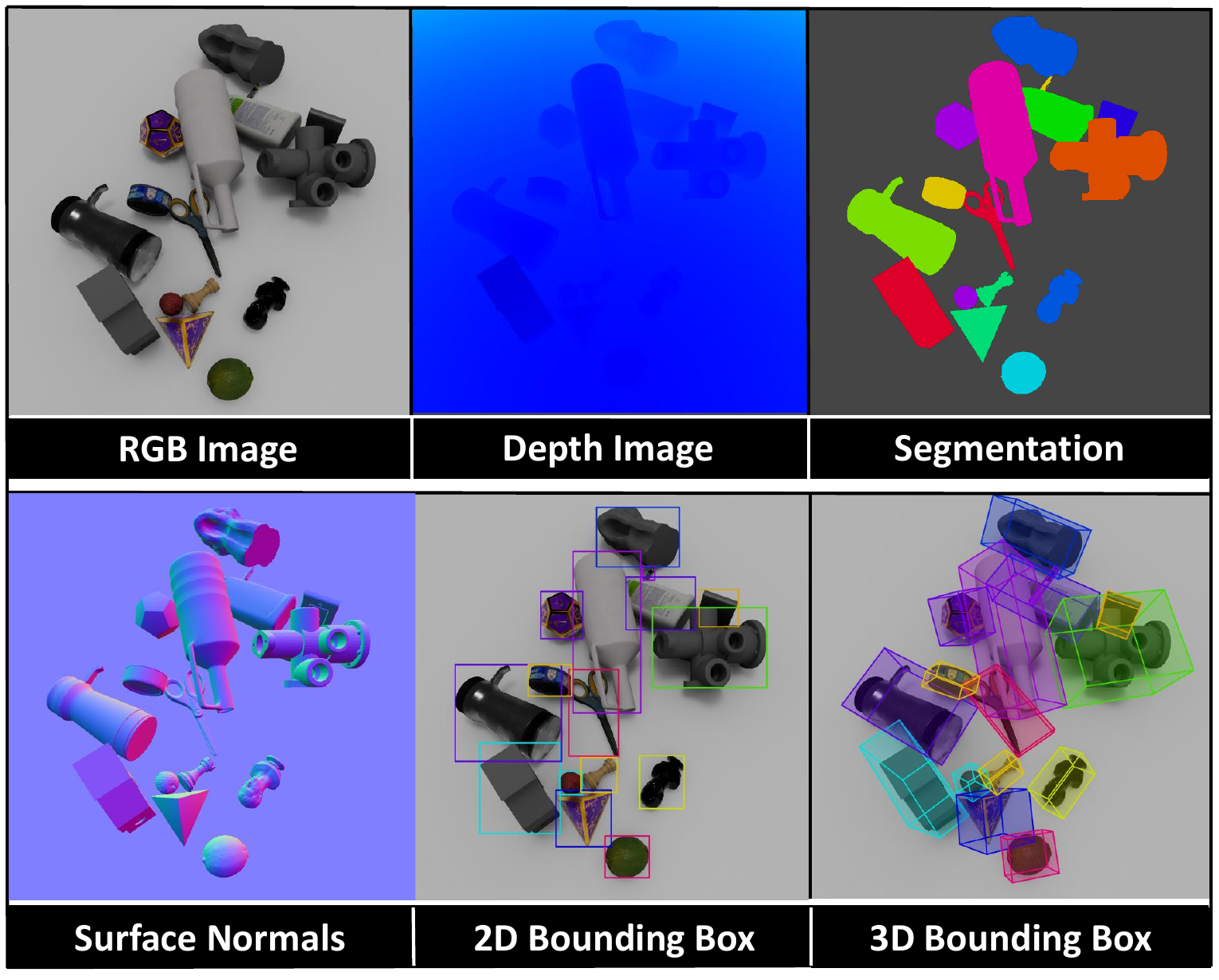}
\vspace{-0.10in}
\caption{\footnotesize 
The Photo-realistic RGB-D images are rendered from a synthetic camera with an intrinsic matrix sampled around the nominal values of a PrimeSense Carmine $1.09$ camera. The segmentation mask is generated using the GPU-RayTracing in PhysX engine~\cite{tracing}. The segmented point cloud can be registered from 2D RGB-D images and surface normals with the help of a segmentation mask using camera intrinsic and extrinsic matrices.
We also provide the 2D and 3D bounding box labels for each object instance, which can contribute to the object detection and pose estimation community.}
\label{dataset}

\end{figure}

\subsection{Cluttered Scene Generation}

The first step of generating the \textit{Sim-Suction-Dataset} is to create random unstructured scenes that contain objects $\mathcal{O}$. In the real world, humans just arbitrarily dump objects $\mathcal{O}$ on a ground plane to create such scenes. We adopt the same strategy in Isaac Sim Simulator~\cite{isaac}. We define our state distribution $\zeta_{scene}$ as a product of the following:
\begin{itemize}
  \item Total objects number $\mathrm{I}$ in one scene: Randomly selected from a range $[1,20]$.
  \item Objects $\mathcal{O}$: uniformly randomly sampled with replacement of total size $\mathrm{I}$ from 20 selected 3D models from YCB-dataset. Objects $\mathcal{O}$ drop locations are uniformly sampled from the 3D space $[-0.1~m,-0.1~m,0.5~m] \times [0.1~m,0.1~m,0.8~m]$ above the ground plane. Objects $\mathcal{O}$'s orientations are uniformly randomly sampled. Each $\mathcal{O}_i$ is dropped at free fall with mass $\mathrm{m}_i$.
  \item Coulomb friction coefficient $\mu$: Randomly sampled from the range $[0,1]$ and used to model tangential forces between contact surfaces.
 \end{itemize} 
 
  For the simulator to implement multi-body physics, we use convex decomposition to describe the collision geometry, where several convex shapes approximate the input mesh. We first sample the initial state from the state distribution $\zeta_{scene}$ to begin the data generation process. Then we start the dynamics simulation and drop objects one-by-one with a rendering time step to avoid object penetration until all objects on the ground plane reach their static equilibrium. Finally, we repeat this process to populate the dataset with $500$ cluttered scenes.

\subsection{Object-aware Suction Grasp Sampling}
Previous sampling methods for cluttered environments use an object-agnostic sampling strategy, where the sampling algorithm searches the point clouds in the entire scene. This method causes a low sampling accuracy and a time-consuming sampling process. It also has poor grasp candidate coverage among different sizes of objects. Additionally, the object-agnostic sampling strategy cannot associate the sampled grasp candidates with object instances and object poses. It considers the point cloud of the cluttered scene as a whole and thus prevents further evaluation. Although the state-of-art semantic segmentation and $6$D pose estimation algorithms can play a part in retrieving object information, we question the estimation accuracy as ``ground truth". To overcome these limitations, we propose an object-aware sampling strategy, which combines the suction grasp candidates with their object information. Owing to the GPU-RayTracing technology in PhysX Engine, we can easily extract each instance's 6D pose and segmentation information from the cluttered environment and reduce the sampling algorithm searching space to each object's point cloud rather than the entire scene. It is commonly assumed that all possible suction poses are infinite in $\mathrm{SE(3)}$ for a single object. Thus, it is impossible for the sampler to cover all of them. Our suction grasp sampling process aims to find a large set of suction candidates evenly distributed on the object surface by considering the suction pad diameter and the time cost for annotating.

\subsubsection{\textbf{Cluttered Scene Point Cloud Processing}}
Our suction grasp sampling process relies on the complete geometries of the object $\mathcal{O}$ in the cluttered scene. In order to get a good geometric description of the cluttered environment due to the existence of object occlusions, we use $800$ synthetic cameras to register segmentation point clouds in $\mathrm{SE(3)}$ and merge the multi-viewed point clouds into a single point cloud $\mathcal{P}$ representing the cluttered scene, then we calculate the surface normals of each point to get local geometry. We save individual object point clouds and cluttered environment point clouds for each scene.

\subsubsection{\textbf{Geometry Guided Approach-Based Sampler}}
given the point cloud of each object instances $i$ from a cluttered scene, we aim to find a set of suction grasp candidates as $\mathbb{S}$=($\mathbb{R}$,$\mathbb{T}$) $\in$ SE(3), which describes the pose and orientation of the suction gripper. To form a seal, we want to align the suction gripper approaching vector with the objects' surface normal on a sampled suction point $t\in\mathbb{T}$. We use iterative Farthest Point Sampling (FPS)~\cite{FPS} to choose a set of points $\mathbb{T}$ from each object point cloud. It is considered to have better coverage on the object surface over the random sampling process. For each sampled suction point $t$ on the differentiable object surface, we calculate the corresponding Darboux frame as $\mathrm{R}\in \mathbb{R}$. A Darboux frame is a natural moving frame constructed on a surface to study curves:
\begin{equation}
R(t)=[v_1(t)|v_2(t)|v_3(t)],
\label{equ:R(t)}
\end{equation}
where $v_1(t)\in \mathcal{N}$ is the normal vector, $v_2(t)$ is the major axis of curvature vector, and $v_3(t)$ is the minor axis of curvature vector. We calculate $v_1(t)$,$v_2(t)$ and $v_3(t)$ by evaluating the Eigenvectors of the $3\times3$ matrix $N(t)$:
\begin{equation}
N(t)=\sum_{t\in\mathbb{T}}^{}\hat{n}(t)\hat{n}^T(t),
\end{equation}
where $\hat{n}(t)$ is the normal vector at point t. 
$[v_3(t),v_2(t),v_1(t)]$ is the Eigenvectors of matrix $N(t)$ in decreasing order. We only align our suction grasp candidates' $X$-axis (Fig.~\ref{ray}) with $v_1(t)$ to ensure that the suction cup makes full contact with the object's surface. The $v_2(t)$ and $v_3(t)$ of the Darboux frame (tangent and binormal vectors) might not be as crucial for suction cup grippers as they are for parallel-jaw grippers, but they can still provide additional information about the local geometry of the object's surface that is useful in seal evaluation and grasp planning.

\subsection{Suction Grasp Candidates Evaluation}
Robots with suction cup grippers interacting physically with the objects in cluttered environments face inherent uncertainties in how objects will react to suction. Previous methods fail to consider the correlation between the object of interest and its surroundings and only evaluate grasp candidates on the singulated object. We believe such methods decrease the accuracy of ground truth labeling by introducing False Positives. We propose a new suction grasp candidates evaluation pipeline, which evaluates the entire cluttered environments with the support of PhysX Engine to provide accurate annotations. The pipeline has three ordered sub-evaluation sequences with binary-valued metric on each suction candidate $\mathrm{S}$ : Collision check $\mathcal{Q}_{collision}(\mathrm{S})=\{0,1\}$, Seal formation evaluation $\mathcal{Q}_{seal}(\mathrm{S})=\{0,1\}$, and dynamics evaluation $\mathcal{Q}_{dynamics}(\mathrm{S})=\{0,1\}$. We define the final metric as a product of the sub-evaluation metrics $\mathcal{Q}(\mathrm{S})= \mathcal{Q}_{collision}(\mathrm{S})\times\mathcal{Q}_{seal}(\mathrm{S})\times\mathcal{Q}_{dynamics}(\mathrm{S})$. 
More detail about each sub-evaluation system is described next:  


\begin{figure}
\centering
\includegraphics[width=1\linewidth]{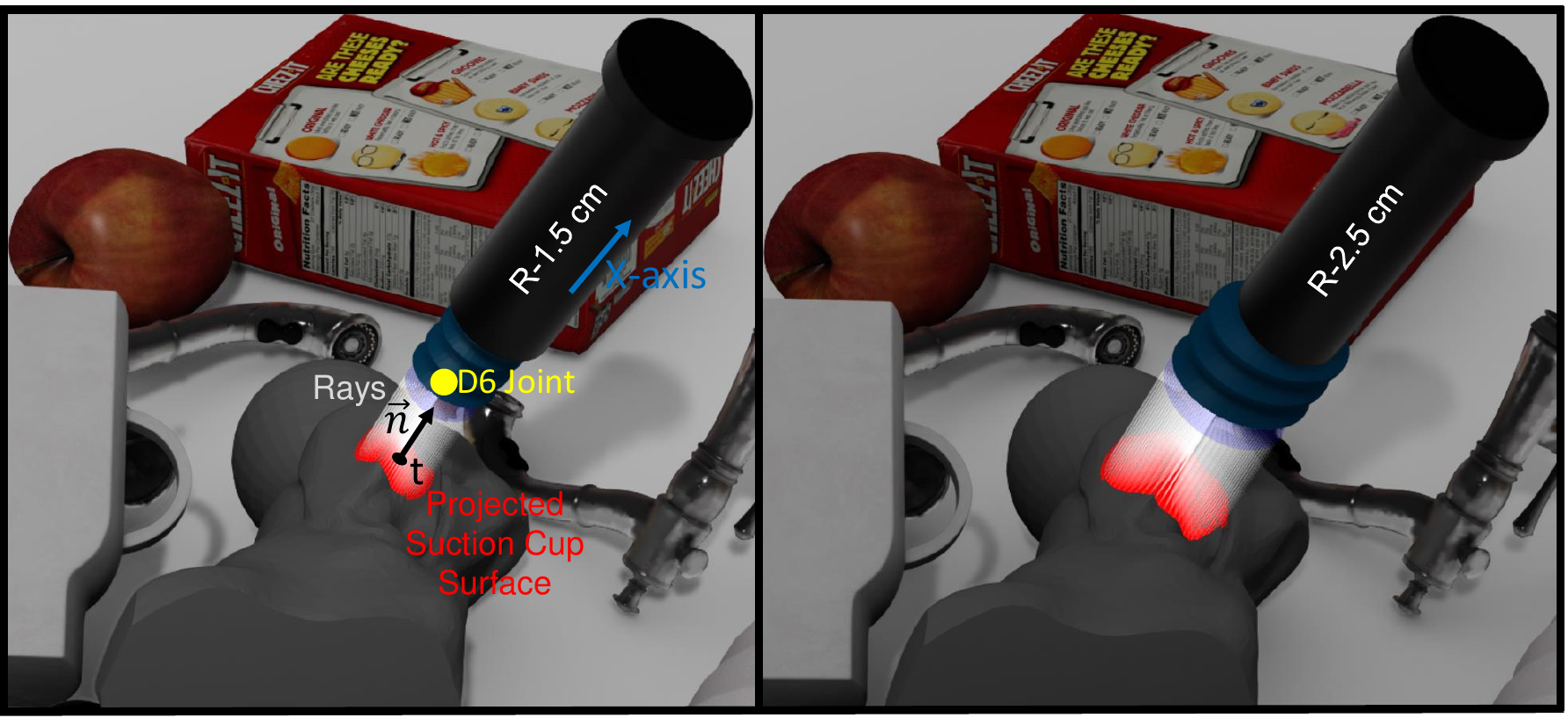}
\vspace{-0.10in}
\caption{\footnotesize \textbf{Left (1.5 cm radius bellows suction cup).} We evaluate the seal performance by casting dense rays along surface normal vectors from the suction cup surface towards the object surface. To evaluate the suction dynamics, we model the suction cup gripper with a 6 degree of freedom joint. We set the suction cup bending angle limit to lock individual axes. We set $20$ N force limit for $1.5$ cm suction cup and check if the 6D joint can be created and maintained during the manipulator movement. \textbf{Right (2.5 cm radius bellows suction cup).} We set the $30$ N force limit for $2.5$ cm suction cup.}
\label{ray}
\vspace{-0.10in}
\end{figure}

\subsubsection{\textbf{Collision Check and Seal Evaluation}}
\begin{figure*}
  \centering
  \includegraphics[width=\linewidth]{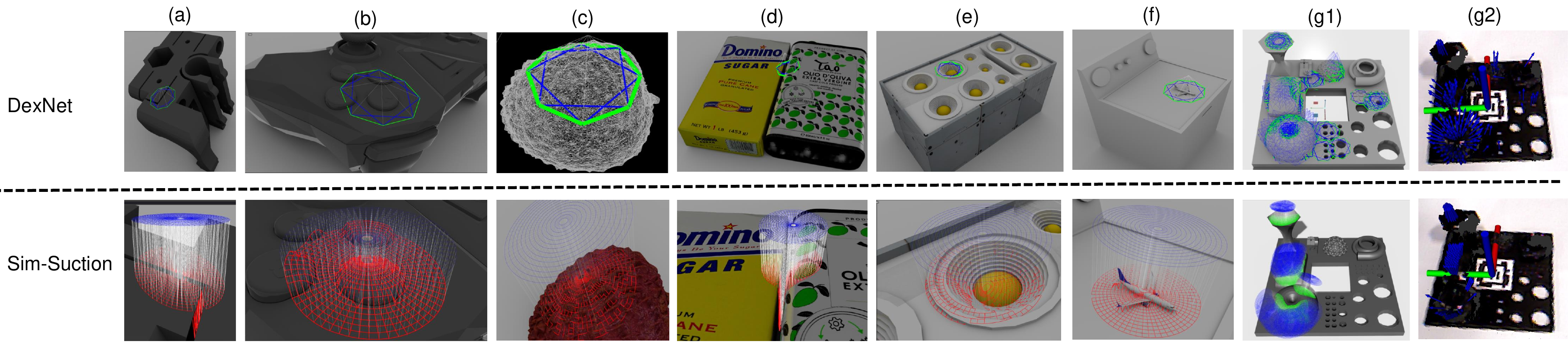}
  \caption{Comparison for 1.5 cm suction cup gripper model. (a) Surface with grooves and holes. (b) Protruding parts. (c) Rough surface. (d) Objects next to each other. (e) Concave surfaces. (f) Overlapped objects. (g1) Customized complex testing board. (g2) 3D printed testing board.}
  \label{model_compare}
\end{figure*}

Suction cup grippers can lift an object when the pressure difference between the atmosphere and the vacuum is large enough. Intuitively, a suction cup
gripper can easily form an airtight seal on a flat surface.
However, forming a seal with the suction cup becomes a
challenge when dealing with an irregular surface. To address
this issue, we took inspiration from a spider’s web and model
the bellows suction cup with 15 concentric polygons, each
with 64 vertices. We perform the collision check in a physics
simulator by casting rays along the x-axis from each vertex,
as shown in the Fig.~\ref{ray}, to detect the closest object that
intersects with a specified ray. We evaluate the suction cup’s
seal by modeling it with deformable material and as a spring,
with a deformable threshold of 10\%. In Fig.~\ref{dataset_study}. (f), the negative correlation shows that the candidate seal evaluation passing rate decreases with increasing object geometry complexity. The trend indicates that it is more difficult for a suction cup gripper to form a seal on a complex object surface. Sim-Suction with a 960-vertex suction cup model can be utilized for assessing suction seals on
intricate geometries, including uneven surfaces and surfaces with holes or grooves. 
We provide a comprehensive comparison of corner cases as shown in Fig.~\ref{model_compare}. In unstructured environments with different difficulty levels of objects, the ground truth labeling process is expected to handle different scenarios and provide an accurate result. However, the previous suction model proposed by Dex-Net and used by others~\cite{suctionnet, zhang} has limitations in dealing with complex geometries and overlapped environments. The DexNet model utilizes the perimeter, flexion, and cone spring connected by eight vertices to assess the seal formation. These vertices are selected on the outer perimeter, resulting in the neglect of any geometry inside the suction cup perimeter. Theoretically, if the suction cup gripper's radius is small enough, and the object is non-porous, there would be no need to be concerned about any geometry inside the suction cup causing air leaks. However, in reality, suction cup grippers are usually larger than the small features commonly found on objects rendering them impossible to ignore. Fig.~\ref{model_compare} (a) shows that a false positive when all eight vertices of the DexNet model sit on a flat surface and the spring deformations are within the threshold, but there is a groove under the suction cup gripper. Fig.~\ref{model_compare} (b) and (e) show that there are geometries under the suction cup gripper that cause the suction cup to deform and not create proper seal.  Fig.~\ref{model_compare} (c) shows that the DexNet model resolution is not suitable for rough surfaces. Fig.~\ref{model_compare} (d) and (f) show that the DexNet analytical model only takes in the singulated object information. In cluttered  environments, DexNet fails to identify neighboring objects, and its collision check performed in DexNet cannot handle these scenarios without object segmentation information. To better quantify our suction model performance, we design and print a 1:1 digital-twin testing board (Fig.~\ref{model_compare}. (g)) consisting of various challenging features such as holes, rough surfaces, and complex geometries. We perform validation experiments with the candidates $\mathcal{Q}_{seal}(\mathrm{S})=1$. The results from Table.~\ref{table:suction_model_compare} show that Sim-Suction suction Model provides more accurate annotations compared to the DexNet suction model. The failure cases in the Sim-Suction experiments are primarily caused by the Apriltag~\cite{apriltag} precision error and imperfections in 3D printing.

\begin{table}
  \centering
  \caption{Comparison of Suction Gripper Analytical Models On Testing Board}
  \label{table:suction_model_compare}
  \small 
  \begin{tabular}{>{\centering\arraybackslash}p{2.5cm} >{\centering\arraybackslash}p{1.5cm} >{\centering\arraybackslash}p{2cm} >{\centering\arraybackslash}p{1.5cm}}
    \toprule
    \textbf{Model} & \textbf{Total Grasps} & \textbf{Successful Grasps} & \textbf{Success Rate} \\
    \midrule
    DexNet & 136 & 83 & 61.03\% \\
    \midrule
    Sim-Suction & 160 & 155 & 96.88\% \\
    \bottomrule
  \end{tabular}
  \begin{tablenotes}

\item This table compares the performance of the DexNet and Sim-Suction suction gripper analytical models on a specially designed testing board consisting of various challenging features. The success rate indicates the percentage of successful grasps out of the total attempted grasps for each model.
\end{tablenotes}
\end{table}

\subsubsection{\textbf{Dynamics Evaluation}}

Previous methods primarily focus on analyzing the external wrench acting on a singulated object. In cluttered environments where the object of interest is at the bottom of a pile, the suction cup gripper must resist not only the wrench due to the object of interest's gravity but also that of the objects above it. Moreover, these methods neglect the dynamics between the suction gripper and the objects. Eppner  \textit{et al.}~\cite{billion} demonstrated that simulations considering the entire grasp process, including dynamics, yield more information about grasp success compared to analytical quality evaluations that only measure static contact quality.
To achieve a more accurate $\mathcal{Q}(\mathrm{S})$, we employ a 1.5 cm radius suction cup mounted on a 7-DoF UR10 robot to simulate the dynamics $\mathcal{Q}_{dynamics}$ using the GPU-enabled Isaac Sim simulator. This simulator utilizes reduced coordinate articulations with Temporal Gauss Seidel (TGS)~\cite{TGS} to compute the future states of objects and the suction cup. We use Riemannian Motion Policy (RMP)~\cite{RMP} to control the UR10 manipulator to reach suction pose configuration $\mathrm{S}$=($\mathrm{R}$,$\mathrm{T}$). 
We model the suction cup gripper as a D6 joint, which represents a 6-degrees-of-freedom constraint defining the relationship between the gripper and the object being grasped. The joint constrains the relative position and orientation of the suction cup and the object, permitting them to function as a single entity. The Temporal Gauss-Seidel (TGS) method is an iterative solver used in physics simulations to compute the future states of objects and the suction cup gripper by solving constraint-based systems efficiently. We calculate $\mathcal{Q}_{dynamics}(\mathrm{S})$ by determining whether the suction cup can create and maintain a 6D joint with suction candidate $\mathrm{S}$ between the object surface and the suction cup surface in a dynamic environment. This process considers various parameters, including force limit, torque limit, friction coefficient, object mass, bellows suction cup maximum bending angle, stiffness, and damping rate. These parameters are essential for TGS to accurately simulate the dynamics and performance of the suction cup gripper modeled as a D6 joint in diverse scenarios, assessing its effectiveness in grasping objects in cluttered environments.We set the force limit and torque limit using data obtained from the silicone 1.5 cm suction cup gripper. These limits represent the maximum force and torque that the gripper can hold before breaking the constraint, affecting the stability and strength of the D6 joint. TGS uses these parameters to decide if the joint can withstand the forces acting upon it during the simulation. The bend angle defines the maximum angle that the suction cup gripper can bend when a load is applied. It helps TGS simulate the deformation of the suction cup when subjected to forces and torques, ensuring that the gripper can maintain a seal with the object surface. The bending stiffness represents the resistance of the suction cup gripper to deformation. TGS uses this parameter to compute the forces and torques required to maintain the shape of the suction cup and ensure proper contact with the object surface. The bend damping parameter helps TGS simulate the energy dissipation in the suction cup gripper during deformation. It contributes to the overall stability and realism of the simulation, particularly when the gripper experiences dynamic forces and torques. By accounting for these parameters, the TGS solver can accurately simulate the dynamics and performance of the suction cup gripper modeled as a D6 joint in various scenarios. We conduct the GPU-based multi-task simulation by trying to lift the objects $\mathcal{O}$ with suction cup configurations $\mathbb{S}$ after passing collision and seal evaluation. In Fig.~\ref{dataset_study}. (g), the negative correlation shows that the candidate dynamics simulation evaluation passing rate decreases with increasing object mass. The trend indicates that it is more difficult for a suction cup gripper to lift a heavy object from various directions.



\section{Suction grasp estimation network}
In this section, we describe our object-aware suction grasp pose estimation network in detail, \textit{Sim-Suction-Pointnet}.

\subsection{Dataset Preprocessing}
The point-wise affordance networks require a binary pixel mask. Given suction grasp candidates $\mathbb{S}=(\mathbb{R},\mathbb{T})$ with $\mathcal{Q}(\mathbb{S})=1$ for each cluttered environment, we use a ball query algorithm on a complete point cloud $\mathcal{P}$ to find all points $\mathcal{P}_t$ that are within a radius of $1.5~cm$ to the query point $\mathrm{T} \in \mathbb{T}$ to represent the contact points between the suction cup and the object surface, and annotate point set $\mathcal{P}_t$ with binary score $1$, and the complement point set $\mathcal{P'}_t$ with binary score $0$. We use the annotated score for each point as the binary point mask for \textit{Sim-Suction-Pointnet}.

\subsection{Sim-Suction-Pointnet Framework}
\label{sec:pointnet}

\subsubsection{\textbf{Affordance Network}}
\begin{figure*}
\centering
\includegraphics[width=0.99\linewidth]{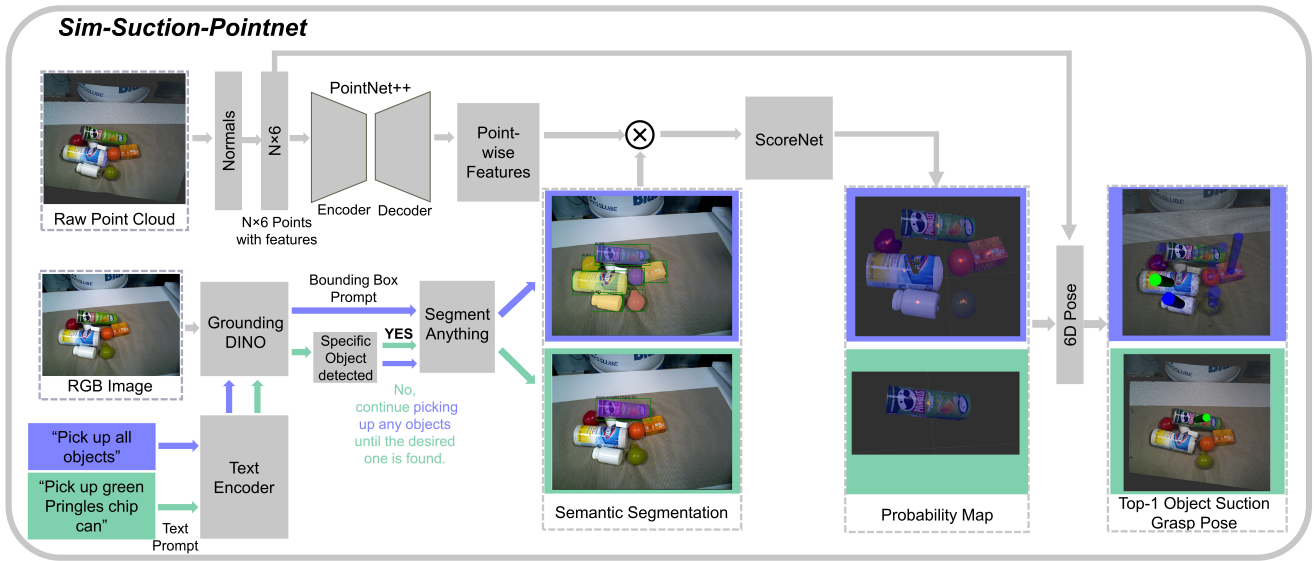}
\vspace{-0.20in}
\caption{\footnotesize The {Sim-Suction} 6D suction grasp pose policy. The green marker represents the 6D grasp pose for the object instance with the highest confidence score. The transparency of the blue markers indicates the confidence score, with higher transparency implying lower confidence and vice versa.}

\label{sim-suction-pointnet}

\end{figure*}

The framework for \textit{Sim-Suction-Pointnet} showed in Fig.~\ref{sim-suction-pointnet}. The \textit{Sim-Suction-Pointnet} is to learn object-aware suction affordance grasping policy in 3D space. We use PointNet++ as our backbone network. The PointNet++ takes raw point clouds $\mathcal{P}$ into the sampling layer, which uses the farthest point sampling (FPS) to choose and normalize a subset of points containing N points with $d$-dim coordinates. The normals layer takes an $N \times d$ matrix as input and outputs an $N \times (d+M)$ matrix, where $M$ is additional point feature channel. We use surface normals for $M$ suggested in ~\cite{pointnet++} as it can increase semantic segmentation performance. We modify the PointNet++ network with the parameters are shown:
\\
$SA(5120,0.02,[128, 128, 256])\rightarrow 
SA(1024,0.08,[256, 256, 512])\rightarrow 
SA(256,0.2,[512, 512, 1024])\rightarrow 
FP(1024, 1024)\rightarrow FP(512,512)
\rightarrow FP(256,256,256)$, \\
where $SA(K,r,[l_1,...,l_d])$ is a set abstraction (SA) level with $K$ local regions of ball radius $r$ using PointNet of $d$ fully connected layers with width $l_i (i=1,..,d)$; $FP (l_1,...l_d)$ is a feature propagation (FP) level with $d$ fully connected layers. 
The decoder of PointNet++ is to turn the group features into point-wise features. 
The PointNet++ loss, $\mathcal{L}_{score}$, is based on MSE loss:
\begin{equation}
   \mathcal{L}_{score}=\frac{1}{N}\sum_{p\in \mathcal{P}}^{}(\mathcal{Q}_p-\hat{\mathcal{Q}_p})^2,
\end{equation}
where $\mathcal{Q}_p$ is the ground truth point score of point $p$, and $\hat{\mathcal{Q}_p}$ is the predicted probability score of point $p$. We trained the network on complete point clouds from $500$ cluttered scenes with a learning rate of $0.001$. To augment the dataset during training, we uniformly randomly select points as centroids and choose $10,000$ nearest points around each centroid. We further augment the dataset by jigging with small rotation angles and scaling to different sizes resulting to $813,451$ point clouds.

\subsubsection{\textbf{Object Detection and Segmentation Mask} }
In complex settings, the depth sensors are noisy. The segmentation models trained on RGB have been shown to produce accurate semantic masks~\cite{noise}. For Sim-Suction-Pointnet, we use the synergy of the point cloud for generating point-wise affordance and RGB image for generating object semantic mask. We add zero-shot Grounding DINO~\cite{groundingdino} as an object detector fine-tuned with \textit{Sim-Suction-Dataset} that takes text prompt as input to generate object bounding boxes, and zero-shot Segment Anything (SAM)~\cite{SAM} that takes bounding boxes as prompt to generate semantic segmentation mask. Zero-shot object detection and segmentation mask methods offer significant benefits when dealing with the challenges of recognizing and segmenting objects from diverse categories without any prior training examples that can transfer knowledge to unseen classes. 

\subsubsection{\textbf{ScoreNet}} 

We integrate a multilayer perceptron (MLP) and output head into our approach to regress and smooth the extracted point-wise features into $N \times 1$ suction probability scores. Utilizing the instance segmentation masks from SAM, we identify object boundaries and filter out suction poses that may result in collisions with other objects in the scene by analyzing the segmentation masks of neighboring objects. We calculate the distance between the centroid or bounding box of the target object and those of other objects in the scene, determining safety margins around each object. These safety margins represent the minimum distance the suction cup should maintain from the object's boundary to avoid collisions. We employ the Darboux Frame to generate 6D suction grasp poses. If a suction pose candidate is found to overlap with the safety margin of any neighboring object, we remove that candidate from the list of potential suction poses, ensuring that the remaining suction poses are collision-free with respect to neighboring objects. Finally, we rank the refined instance suction pose candidates based on their updated suction grasp affordance scores.

\begin{figure*}
\centering
\includegraphics[width=0.9\linewidth]{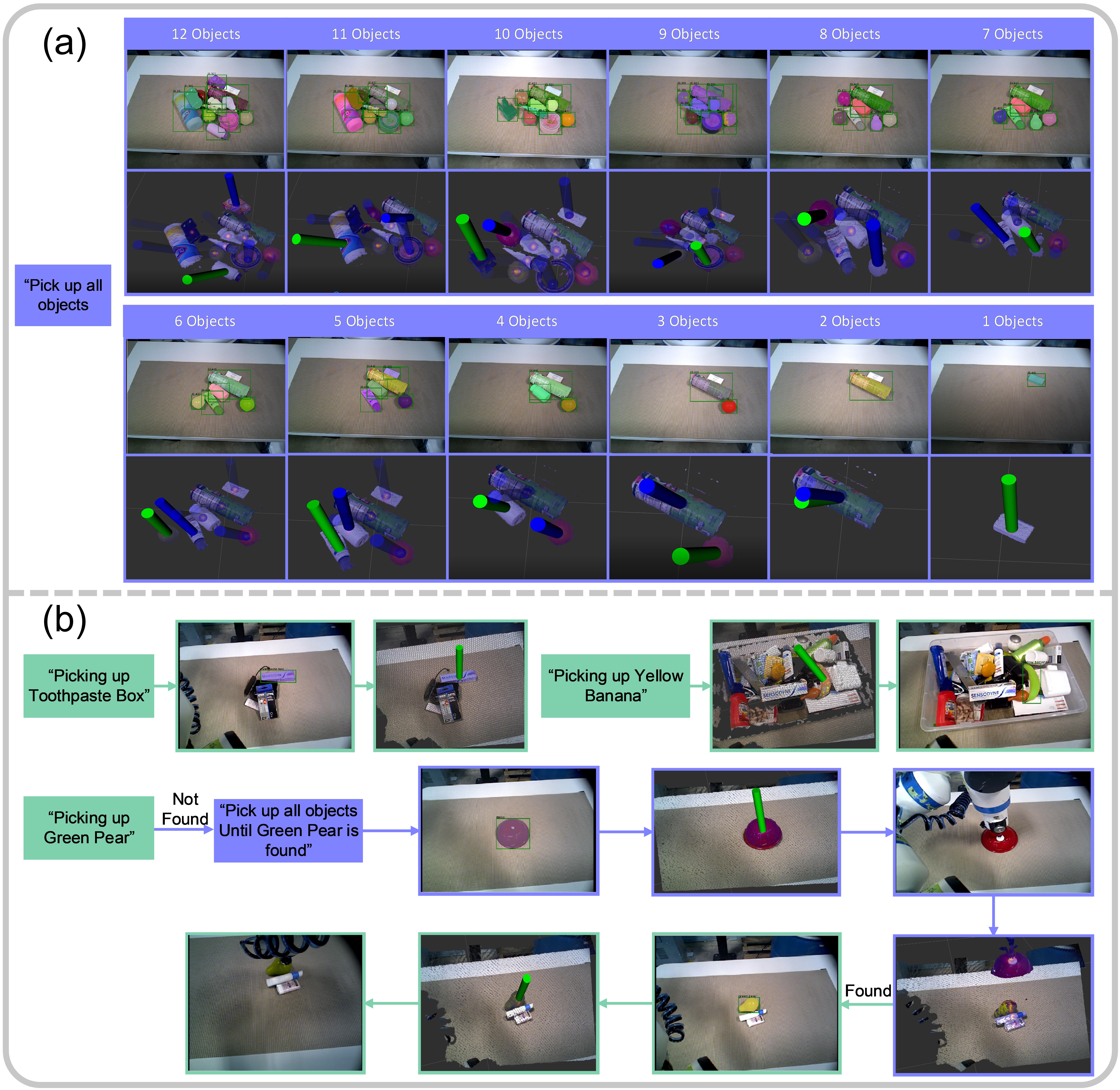}

\caption{\footnotesize The \textit{Sim-Suction} policy task sequence examples. The policy demonstrates robust grasping reliability in real-world scenarios. The figure displays the policy applied in two tasks: (a) "pick up all objects", where the robot continuously attempts grasps until the table surface is clear, and (b) "pick up a specific object", where the policy focuses on grasping a target object based on the text prompt input.} 
\label{exp}
\vspace{-0.1in}

\end{figure*}

\label{network}

\section{EXPERIMENTS}
\begin{figure}
\centering
\includegraphics[width=1\linewidth]{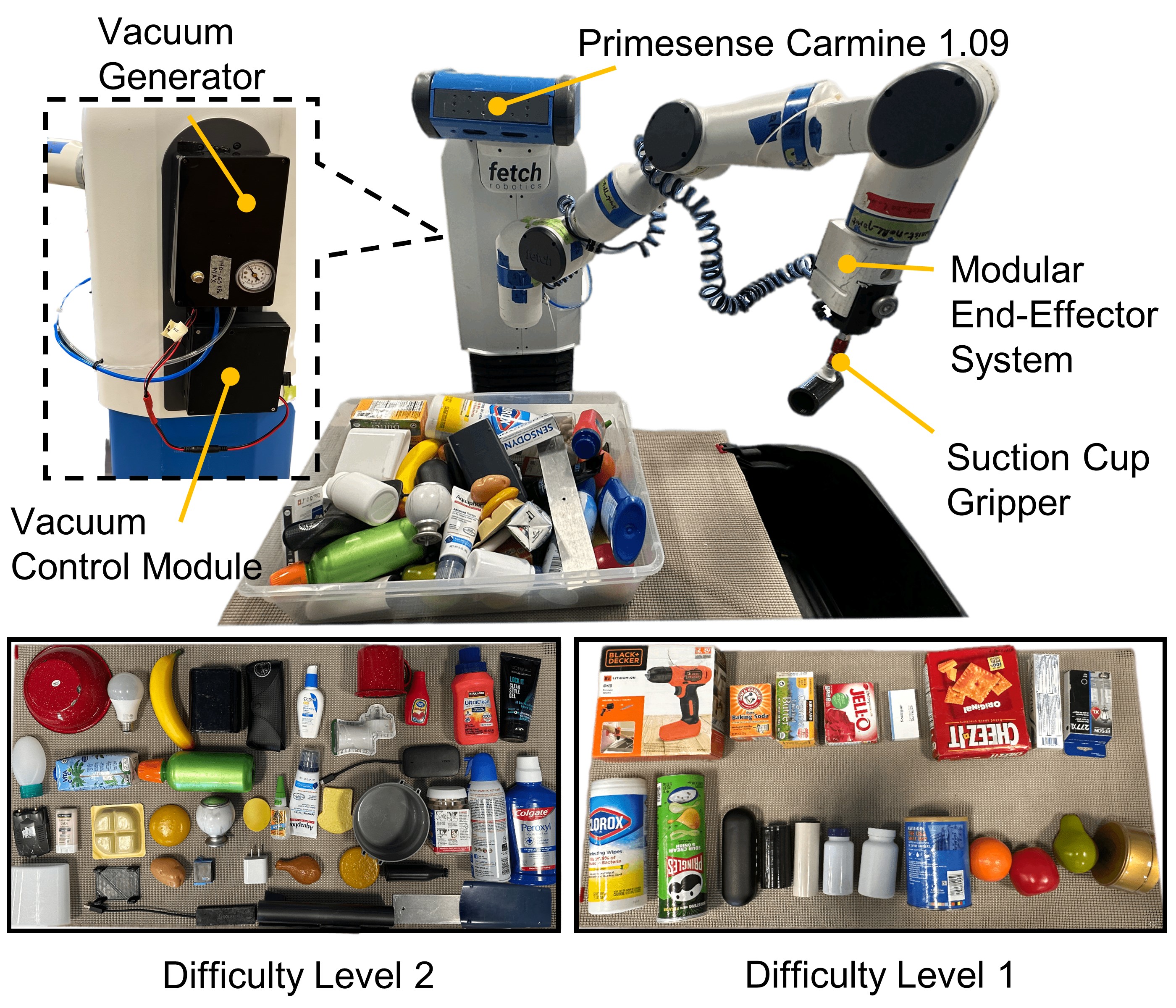}
\vspace{-0.10in}
\caption{\footnotesize \textbf{(Top)} The experimental setup with a Fetch robot equipped with the Modular End-Effector System\cite{modular_end_effector}. \textbf{(Bottom)} We choose 60 household items, with 20 objects in Level 1 (primitive shapes) and 40 objects in Level 2 (varied geometries). These objects are considered novel to the \textit{Sim-Suction-Pointnet} policy, as it has no prior knowledge of them. The objects feature a range of challenging characteristics, such as complex geometries, irregular shapes, and varied surface textures, making the task more difficult.}
\label{exp_setup}
\vspace{-0.15in}

\end{figure}

In this section, we first utilize an online evaluation system to explore the effects of dataset diversity and point cloud type, comparing the results with baselines on both similar and novel datasets. Subsequently, we perform an ablation study with real robot experiments to investigate the impact of the segmentation mask, contrasting our findings with baseline approaches. Lastly, we conduct extensive real robot experiments to assess the suction grasp success rate of our \textit{Sim-Suction} policy when retrieving novel objects from a variety of cluttered environments and compare it with state-of-the-art methods.

\subsection{Online Ablation Study}
\label{baseline}

Previous methods employ an offline evaluation system to calculate the Average Precision (AP) by comparing the inferred affordance score with the pre-annotated ground truth from the dataset. However, pre-annotated ground truth cannot cover all possible suction poses that exist, leading to inaccurate AP results. To address this issue, we adopt the online evaluation system from Section IV.D. Given a set of 6D suction configurations $\mathbb{S}=(\mathbb{R},\mathbb{T})$ and the corresponding confidence scores after inference, we consider a suction pose $\mathrm{S}\in\mathbb{S}$ as a true positive if ${Q(\mathrm{S})}=1$, where $\mathcal{Q} =\mathcal{Q}{seal} \times \mathcal{Q}{collision}\times \mathcal{Q}_{dynamics}$. We conduct all performance evaluation experiments on an NVIDIA RTX 3080 Ti GPU.

{
Our baseline method uses the single-viewed point cloud to predict the affordance score by estimating the variance of the surface normals on each point with its nearby neighbors using ball query with a radius of $1.5~cm$. The baseline method aims to calculate the object surface flatness around that point, where high variance means low flatness. We use an instance segmentation mask to remove the ground plane because it is not our region of interest and has the highest flatness score. Sim-Suction-Pixelnet method uses DeepLabV3+ as backbone and trained with Sim-Suction-Dataset that takes RGB-D images as input and outputs a pixel-wise affordance map. We use the same 6D pose layer from \textit{Sim-Suction} to process all affordance scores and output the 6D suction grasp poses. Table~\ref{table:online ablation} presents the performance of various networks under different training sizes and test conditions, comparing their Average Precision (AP) on both similar and novel datasets. Similar objects refer to objects that share common characteristics with those in the training dataset but with different scales. Novel objects are objects that are introduced during the testing phase and are not part of the training dataset. The results demonstrate the importance of dataset diversity, point cloud type, and point-wise learning in training the models for improved grasp prediction.

\subsubsection{\textbf{Dataset Diversity}}

To demonstrate the importance of a large-scale dataset, we evaluate the performance of our method on both a similar dataset and a newly generated novel dataset, which includes 100 unique objects. The \textit{Sim-Suction-PointNet} performance increases with the increase in dataset diversity.

\subsubsection{\textbf{Effect of Point Cloud Type}}

To illustrate the rationale for training on complete point clouds merged by multi-view camera frames for \textit{Sim-Suction-PointNet}, we evaluate the performance and compare it to \textit{Sim-Suction-PointNet} trained with a single-viewed point cloud. The results show that \textit{Sim-Suction-PointNet} achieves slightly better performance across all dataset diversities, even when inference is performed using a single-view camera. One possible reason is that multi-view merged point clouds provide a more complete and detailed representation of the object, capturing its various features and geometries from multiple perspectives. This richer representation enables the model to learn more robust and generalized features during training, leading to better performance during inference.

\subsubsection{\textbf{Effect of Point-Wise Learning}}

Our \textit{Sim-Suction-PointNet}, trained with point clouds, demonstrates better performance on novel objects compared to Sim-Suction-PixelNet, which is trained with RGB-D images. One possible reason is that point clouds directly represent the 3D structure of the scene, providing precise geometric information about the objects. This information enables the model to better understand the shape and size of the objects, which in turn helps it learn more effective grasp affordances for novel objects. Point cloud representations are more invariant to viewpoint and scale changes compared to RGB-D images. This allows the model to generalize better across different object orientations, sizes, and camera viewpoints, leading to improved performance on novel objects.

\begin{table*}
\caption{Online Ablation Study of Networks for Different Training Sizes and Test Conditions}
\label{table:online ablation}
\centering
\begin{tabular}{c|c|cccc|cccc}
\hline
\multirow{2}{*}{Training Size} & \multirow{2}{*}{\centering Network} & \multicolumn{4}{c}{Test Similar} & \multicolumn{4}{c}{Test Novel} \\ \cline{3-10}
 & & Top-1 & Top-1\% & Top-5\% & Top-10\% & Top-1 & Top-1\% & Top-5\% & Top-10\% \\ \hline
\multirow{4}{*}{20 objects, 100 scenes} & Baseline & 66.34 & 64.96 & 60.44 & 52.66 & 65.12 & 64.25 & 59.76 & 51.43 \\
 & Sim-Suction-Pixelnet & 88.04 & 85.85 & 79.87 & 74.77 & 77.01 & 74.27 & 69.91 & 65.36 \\
 & \textit{Sim-Suction-Pointnet (SV-PCL)} & 84.81 & 79.95 & 75.98 & 68.61 & 81.41 & 77.84 & 76.2 & 66.56 \\
 & \textit{Sim-Suction-Pointnet (MV-PCL)} & 85.72 & 81.86 & 77.46 & 71.92 & 83.43 & 80.54 & 77.87 & 70.36 \\ \cline{1-10}
\multirow{3}{*}{\makecell{20 objects, 500 scenes}} & \textit{Sim-Suction-Pointnet (SV-PCL)} & 86.63 & 83.77 & 78.94 & 75.23 & 85.45 & 83.24 & 79.54 & 74.16 \\
 & \textit{Sim-Suction-Pointnet (MV-PCL)} & 87.54 & 85.68 & 80.42 & 78.54 & 87.47 & 85.94 & 81.21 & 77.96 \\ \cline{1-10}
\multirow{2}{*}{\makecell{1550 objects, 500 scenes}} & \textit{Sim-Suction-Pointnet (SV-PCL)} & 88.45 & 87.59 & 81.9 & 81.85 & 89.49 & 88.64 & 82.88 & 81.76 \\
 & \textit{\textbf{Sim-Suction-Pointnet (MV-PCL)}} & \textbf{89.36} & \textbf{89.5} & \textbf{83.38} & \textbf{81.16} & \textbf{91.51} & \textbf{91.34} & \textbf{84.55} & \textbf{82.56} \\ \hline
\end{tabular}
\begin{tablenotes}
      \item Abbreviations: SV-PCL refers to a model trained on single-view point clouds, while MV-PCL refers to a model trained on multi-view merged point clouds. Top-1, Top-1\%, Top-5\%, and Top-10\% represent the performance metrics for different confidence percentiles. 

    \end{tablenotes}
\end{table*}

\begin{figure*}
\centering
\includegraphics[width=1.0\linewidth]{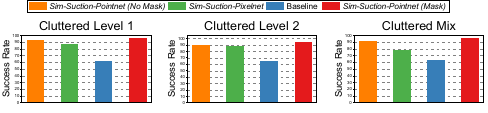}

\caption{\footnotesize This figure presents the ablation study results, showing the success rate of attempted grasps in cluttered environments using different methods. \textbf{Cluttered Level 1 objects}: \textit{Sim-Suction-Pointnet (No Mask)} achieved a success rate of 93.33\% (56 successes in 60 attempts), \textit{Sim-Suction-Pixelnet} 86.89\% (53 successes in 61 attempts), the Baseline method 62.50\% (35 successes in 56 attempts), and \textit{Sim-Suction-Pointnet (Mask)} 96.67\% (58 successes in 60 attempts). \textbf{Cluttered Level 2 objects}: \textit{Sim-Suction-Pointnet (No Mask)} achieved a success rate of 90.00\% (36 successes in 40 attempts), \textit{Sim-Suction-Pixelnet} 88.57\% (31 successes in 35 attempts), the Baseline method 64.52\% (20 successes in 31 attempts), and \textit{Sim-Suction-Pointnet (Mask)} 95.00\% (38 successes in 40 attempts). \textbf{Cluttered Mixed Level 1 and Level 2 objects}: \textit{Sim-Suction-Pointnet (No Mask)} achieved a success rate of 91.84\% (45 successes in 49 attempts), \textit{Sim-Suction-Pixelnet} 78.26\% (36 successes in 46 attempts), the Baseline method 63.64\% (21 successes in 33 attempts), and \textit{Sim-Suction-Pointnet (Mask)} 95.92\% (47 successes in 49 attempts).}
\label{exp_fig}
\end{figure*}
 








\subsection{Real Robot Experiments Setup}
To further evaluate the \textit{Sim-Suction} performance in the real world and address the domain gap problem, we perform experiments with a Fetch mobile manipulation platform equipped with a Primesense Carmine 1.09 head camera and a modular end-effector system~\cite{modular_end_effector} with interchangeable 1.5cm radius suction cups with multi-bellow designs rated for 1.3kg payload (Fig.~\ref{exp_setup}). The inference and grasping planning algorithms run on a remote laptop with an NVIDIA GeForce 3070Ti GPU. The Fetch robot and the vacuum pump control module communicate with the remote laptop via ROS nodes. To initiate the experiments, the Fetch robot approaches the workbench and positions itself to observe the tabletop. Once in its initial position, the robot's base remains static throughout the operation, as base movement is not required for arm movement and grasp planning in our setup. While the base is fixed, the robot's torso is capable of vertical movement to adjust its viewing angle and arm height as needed, which is considered part of the motion planning. As a result, the camera height may vary across trials due to the torso adjustments, creating an arbitrary viewpoint for each experiment and testing the model's adaptability. Given the limitations of our camera setup in capturing fine object details, we made selective decisions about the objects included in the real-world experiments. Specifically, Level 3 objects with intricate details were excluded, as our manipulator's camera resolution was insufficient to accurately capture their nuances, affecting grasp prediction performance. Our focus was primarily on Level 1 and Level 2 objects, as these categories represent most objects commonly encountered. We selected 60 novel objects for our experiments, which the policy had no prior knowledge of. The objects were split into two difficulty levels shown in Fig.~\ref{exp_setup}: Level 1: $20$ objects with only primitive shapes, and Level 2: $40$ objects with varied geometries. For Level 1, since it has fewer objects, we dump all 20 objects onto the table to create a confined environment. For Level 2, we place the 40 objects in a bin. For the cluttered mix, we put all 60 objects in the bin. This experimental setup further tests the performance of the \textit{Sim-Suction} policy in handling objects with different shapes and difficulty levels under various environmental conditions.

\subsection{Experimental Results}

We employ a strict  reliability metric to evaluate the \textit{Sim-Suction} performance on grasping the selected objects: the ratio of the number of successful grasps to the total number of attempts. This metric is stringent as it accounts for every individual attempt, without aggregating any subsequent attempts even if the item is ultimately grasped. This assessment highlights the policy's ability to rapidly and accurately identify suitable grasping points on novel objects, underscoring the significance of robust performance in each grasp attempt. The ultimate test of \textit{Sim-Suction} is to execute the policy in the real world and deal with domain gap. We want to show our \textit{Sim-Suction} policy trained on large-scaled synthetic point cloud data can transfer well to reality and achieve robust grasp reliability. Fig.~\ref{exp} shows the Sim-Suction policy on example tasks. For the "pick up all objects" task (Fig.\ref{exp}. (a)), the robot performs continuous grasp attempts until no objects remain on the table surface. This task is not sensitive to the text prompt input. For the "pick up a specific object" task (Fig.\ref{exp}. (b)), if the object of interest is found, the policy executes the pick-up; otherwise, the policy will first carry out the "pick up all objects" task to search for the object of interest. If found, the policy will pick up the target object and complete the task. This task is highly sensitive to the object description provided by humans as a text prompt. We primarily concentrate on the "pick up all objects" test for several key reasons. First, it enables a thorough assessment of the \textit{Sim-Suction} policy by challenging its adaptability and versatility across a wide range of objects with different shapes, sizes, and geometries. Second, focusing on this task helps evaluate the policy's robustness in terms of continuous performance, providing insights into the model's reliability and efficiency in real-world settings. Moreover, the "pick up all objects" task is less sensitive to the text prompt, which allows us to focus on the core aspects of the grasping policy. We initially conduct an ablation study on a subset of testing objects to investigate the impact of segmentation masks on improving the success rate for the "pick up all objects" task and compare it to the baselines. Then, we conduct comprehensive experiments to pick up all 60 objects via a series of experiments to compare the \textit{Sim-Suction} performance against the current state-of-the-art method DexNet 4.0 Suction (FC-GQ-CNN-4.0-SUCTION). We utilize a robot equipped with the MoveIt! motion planning framework to execute the suction grasp with the highest confidence score, which is represented by a green marker in Fig.\ref{exp}. (a). If the motion planning framework fails to find a valid solution, the policy proceeds with the next-best suction grasp, indicated by a solid blue marker in Fig.\ref{exp}.
If the motion planning framework continues to fail in finding a valid solution, the policy will proceed with the subsequent suction grasp options based on their confidence scores, in descending order. 

\subsubsection{\textbf{Ablation Study (picking up all objects)}}
We conduct ablation experiments on a subset of test objects, which includes 6 objects from level 1 and 8 objects from level 2. We compare \textit{Sim-Suction-Pointnet (Mask)} with \textit{Sim-Suction-Pointnet (No Mask)} and the baselines described in \ref{baseline}. As shown in Fig.~\ref{exp_fig}, our \textit{Sim-Suction-Pointnet (Mask)} achieves the highest success rate across all cluttered categories, with a reliability of $\textbf{96.67\%}$ for cluttered level 1 objects, $\textbf{95.00\%}$ for cluttered level 2 objects, and $\textbf{95.90\%}$ for cluttered mixed objects. \textit{Sim-Suction-Pointnet (Mask)} outperforms \textit{Sim-Suction-Pointnet (No Mask)}, indicating the importance of segmentation masks. These masks provide additional information about instance boundaries, which can be vital in identifying suitable grasping points on the object's surface. When the model has access to this information, it can better focus on the target object and avoid interference from surrounding objects or clutter. The use of segmentation masks also prevents the policy from repeatedly attempting the same unsuccessful suction pose. Instead, it allows the policy to shift its focus to other objects. \textit{Sim-Suction-Pointnet} performs better than \textit{Sim-Suction-Pixelnet}. learning point-wise features from point clouds, which allows the model to focus on local geometric properties and relationships between points. \textit{Sim-Suction-Pointnet} processes raw point cloud data, which is less affected by the domain gap between synthetic and real-world data. On the other hand, \textit{Sim-Suction-Pixelnet} relies on RGB-D images, which are more sensitive to variations in lighting, textures, and other factors that may differ between simulated and real environments. The robot with the Baseline policy takes longer to get prediction results. The executed suction grasp poses from the Baseline policy have collisions with nearby objects in many cases that cause the failure.

\subsubsection{\textbf{Experiments (picking up specific objects)}}
To evaluate the effectiveness of our method in executing tasks that require selecting specific novel objects (Fig.\ref{specific}), we create cluttered environments using a small set of 8 objects. In these scenarios, the object of interest may be either visible to the camera or hidden beneath a pile of other items. The policy's objective is to search for and locate the target object, successfully grasp it, and complete the task. As the method is sensitive to the text prompt, we perform pre-tests and refine the text prompt to generate a reasonable description for the novel objects of interest. We conduct 20 experiments to pick up specific objects, and the policy achieves a success rate of 16/20. The failure cases occur when the robot picks up other objects due to false detection of the novel objects, as these objects are very similar and lack textures. Further discussion on this issue can be found in the Grounding DINO paper\cite{groundingdino}.

\begin{figure}
\centering
\includegraphics[width=1\linewidth]{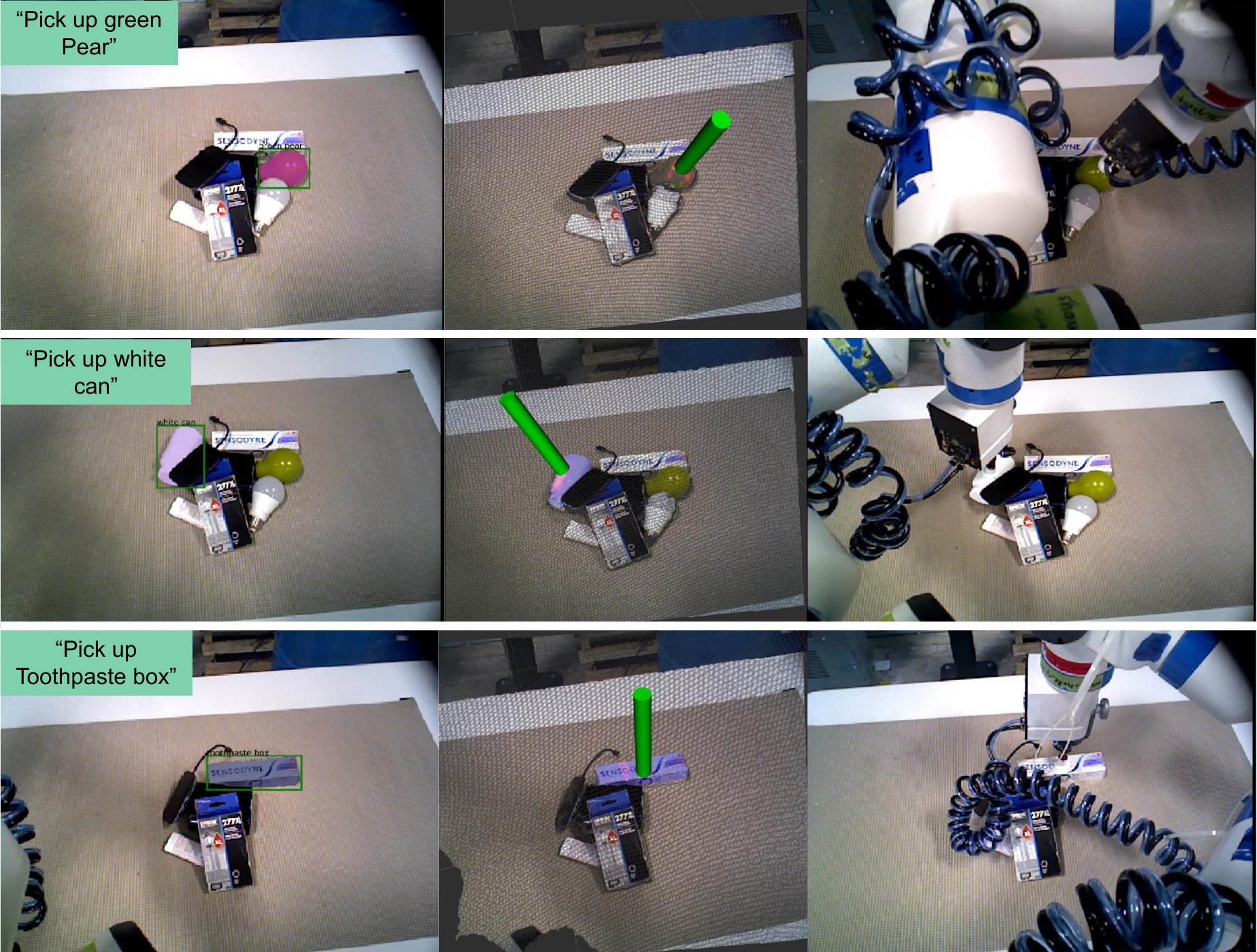}

\caption{\footnotesize Qualitative results of experiments for picking up specific objects. The figure displays various instances where the \textit{Sim-Suction-Pointnet} policy successfully identifies and grasps the target object in cluttered environments.} 
\label{specific}
\end{figure}

\subsubsection{\textbf{Comprehensive Experiments (picking up all objects)}}

\begin{figure}
\vspace{-0.2in}
\centering
\includegraphics[width=1\linewidth]{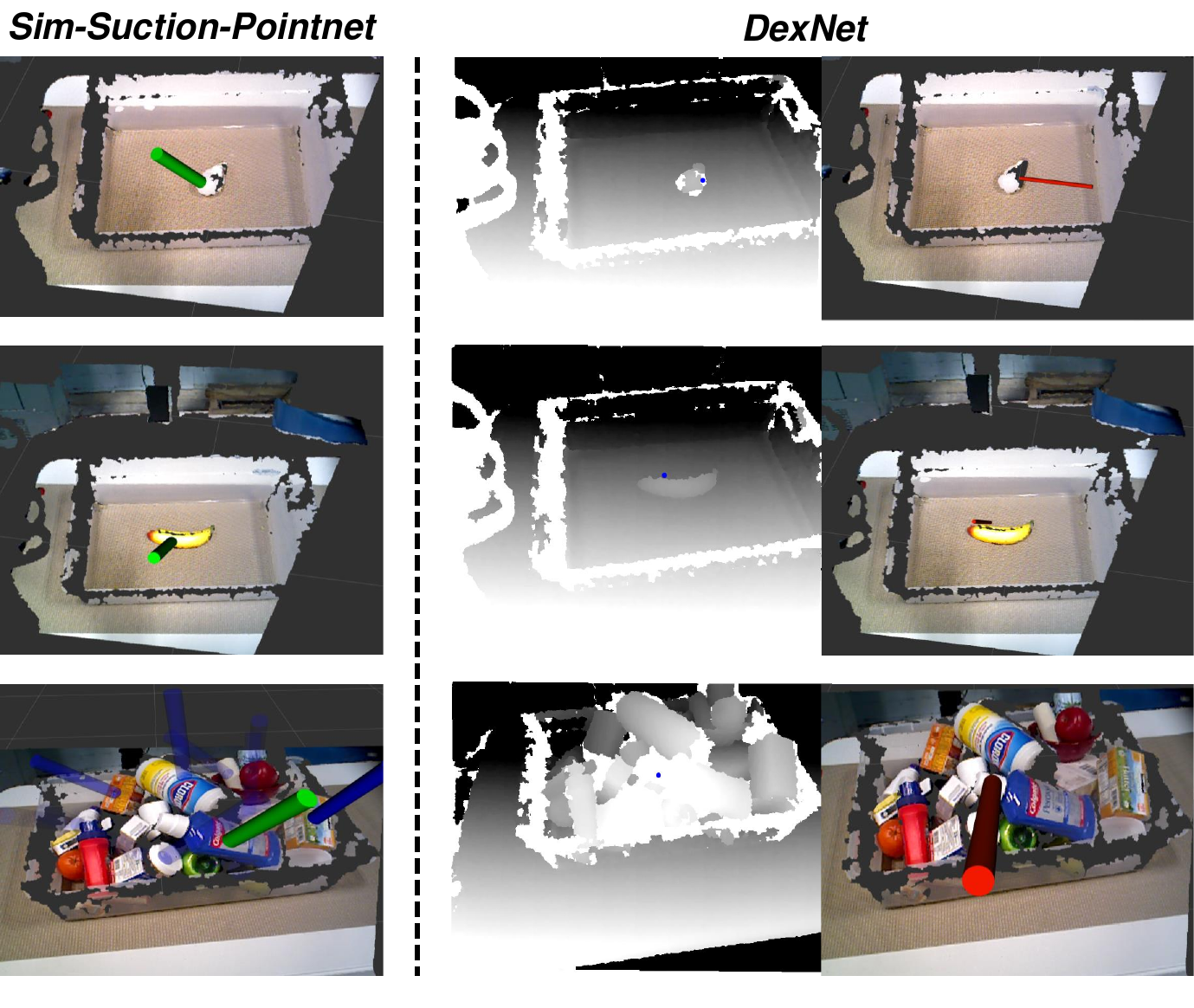}

\caption{\footnotesize Qualitative comparison of \textit{Sim-Suction-Pointnet} \textbf{(Left)} with DexNet \textbf{(Right)}. Top-Row: DexNet generates suction poses on the object edge. Middle-Row: DexNet generates suction poses on the nearby ground. Bottom-Row: DexNet generates suction poses on the unsuctionable area of the object.} 
\vspace{-0.2in}
\label{dex_exp}
\end{figure}

To better evaluate and quantify the reliability of \textit{Sim-Suction} in cluttered environments, we perform around $600$ attempts and compare them with the state-of-the-art DexNet. By comparing our method with the state-of-the-art, we can establish a benchmark for future research in this area. This comparison allows us to measure the progress made by our method and identify areas where further improvements can be made. 
Table~\ref{table:results} showcases the experimental outcomes for successful attempts against total attempts in cluttered environments for various methods. DexNet-4.0 (GQ-CNN) is trained on synthetic depth images captured from a fixed top-down camera view. Both DexNet and our method aim to test on novel objects that are not present in our respective datasets, emphasizing the zero-shot generalization capabilities. In comparison, our experiments employ a mobile robot with a changing camera viewpoint, which considerably diverges from DexNet's training environment. The \textit{Sim-Suction-Pointnet} (Mask) approach demonstrates superior performance across all cluttered environments, with reliability rates of $\textbf{96.76\%}$, $\textbf{94.23\%}$, and $\textbf{92.39\%}$ for cluttered level 1, cluttered level 2, and cluttered mixed objects, respectively. The state-of-the-art DexNet 4.0-Suction exhibits a lower reliability of $81.22\%$, $77.73\%$, and $71.61\%$ for the same scenarios. The results directly reported by DexNet 4.0-Suction~\cite{dexnet4.0} are $93\%$, $80\%$, and $78\%$. In DexNet's experiment setup, they chose 50 objects for a cluttered mix environment, while our setup consists of around 60 objects. DexNet performs worse in our experiment setting. One possible reason for this discrepancy is that the experiments conducted in DexNet 4.0 use an industry-level over-bin depth camera, whereas our experiment employs a changing camera view. DexNet 4.0 is also trained on a fixed vision dataset, resulting in suboptimal performance on mobile manipulation platforms. As mentioned by the authors in a seminar\cite{talk}, Dex-Net 4.0 faces challenges in mobile manipulation platforms with a moving camera that is not mounted on top of the workspace. Figure~\ref{dex_exp} illustrates several instances where DexNet encounters difficulties. These challenges arise due to noisy depth images, leading DexNet to fail in generating reachable 6D suction poses, which are typically located on the object boundary. Additionally, DexNet employs a segmentation method that only separates the foreground of the scene, rather than employing instance segmentation. As a result, it struggles to handle individual objects when they are placed in cluttered environments. The results emphasize the effectiveness and adaptability of the \textit{Sim-Suction-Pointnet} (Mask) method to various camera perspectives and real-world conditions in intricate cluttered environments. In contrast to the depth images used by DexNet-4.0 (GQ-CNN), \textit{Sim-Suction-Pointnet} (Mask) employs a point cloud-based strategy and uses synergy with a zero-shot RGB segmentation method. This enables \textit{Sim-Suction-Pointnet} (Mask) to be more resilient and adaptable to different camera angles and novel objects. By using the segmentation mask, \textit{Sim-Suction-Pointnet} (Mask) refines the point cloud input and separates the object of interest from the surrounding clutter. This focus on the target object enhances the model's ability to pinpoint appropriate grasping points. Furthermore, the extensive synthetic \textit{Sim-Suction-Dataset} utilized for training \textit{Sim-Suction-Pointnet} (Mask) encompasses a wide variety of object shapes, sizes, and geometries, as well as provides more accurate ground truth labeling. This diverse dataset contributes to the policy's superior generalization abilities in comparison to DexNet-4.0 (GQ-CNN). Examples of failure cases encountered by \textit{Sim-Suction-Pointnet} during experiments (Fig.~\ref{fail}) are primarily situations where the object is obscured from the camera's view since we use a moving camera instead of a high-resolution fixed vision system above the bin. Other cases arise from the nature of cluttered environments, where the object is not stable and may move, rotate, or roll during the grasping process. Only a few cases are caused by the unsuccessful attempts to grasp an object beneath a pile.

\begin{table}
\caption{Experimental results of success attempts versus total attempts in cluttered environments for different methods}
\label{table:results}
\centering
 \scalebox{0.70}{
\begin{tabular}{c|c|c|c|c}
\toprule
\textbf{Policy} & \textbf{\# Attempts} & \textbf{\# Total Attempt Fail.} & \textbf{Success Rate} & \textbf{\shortstack{Objects\\ Grasped/Total}} \\
\midrule
\multicolumn{5}{c}{\textit{Cluttered level 1 (20 Objects per Scene)}} \\
\midrule
DexNet-4.0 (GQ-CNN) & 213 & 40 & 81.22\% & 173/179 \\
\textbf{\textit{Sim-Suction-Pointnet} (Mask)} & \textbf{185} & \textbf{6} & \textbf{96.76\%} & \textbf{179/179} \\
\midrule
\multicolumn{5}{c}{\textit{Cluttered level 2 (40 Objects per Scene)}} \\
\midrule
DexNet-4.0 (GQ-CNN) & 238 & 53 & 77.73\% & 185/197 \\
\textbf{\textit{Sim-Suction-Pointnet} (Mask)} & \textbf{208} & \textbf{12} & \textbf{94.23\%} & \textbf{196/197} \\
\midrule
\multicolumn{5}{c}{\textit{Cluttered Mix (60 Objects per Scene)}} \\
\midrule
DexNet-4.0 (GQ-CNN) & 229 & 65& 71.61\% & 164/172 \\
\textbf{\textit{Sim-Suction-Pointnet} (Mask)} & \textbf{184} & \textbf{14} & \textbf{92.39\%} & \textbf{170/172} \\
\bottomrule
\end{tabular}
}

\end{table}

\begin{figure}
\centering
\includegraphics[width=1.0\linewidth]{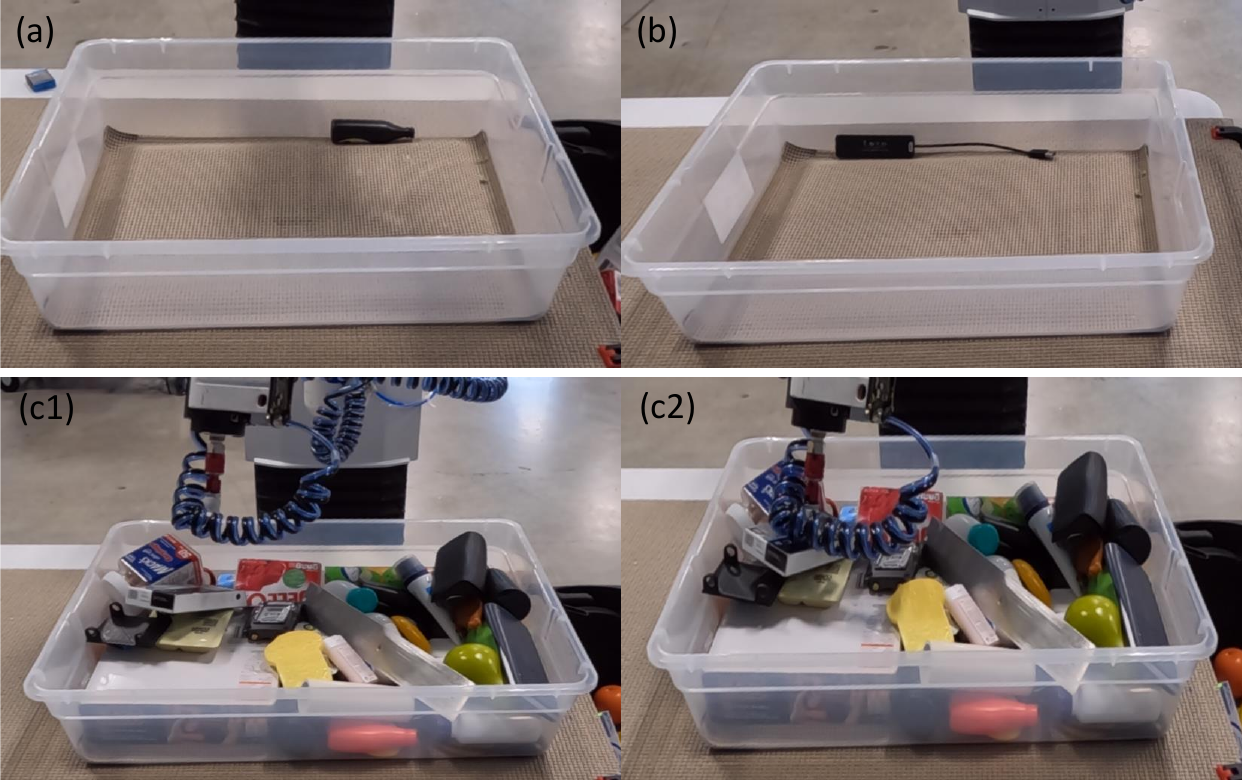}

\caption{\footnotesize Examples of failure cases. (a) and (b) The object overlaps with the bin edges. (c1) and (c2) The object is unstable, causing it to move when the robot attempts to form a seal.}
\label{fail}
\end{figure}
\vspace{-0.1in}

\section{CONCLUSIONS \& FUTURE WORK}
\vspace{-0.05in}
In this paper, we present \textit{Sim-Suction}, a deep learning-based object-aware suction grasp policy for objects in cluttered environments. Experiments conducted on a mobile manipulation platform demonstrate that \textit{Sim-Suction}, learned from the synthetic point cloud dataset \textit{Sim-Suction-Dataset}, achieves a robust success rate in real-world cluttered environments with dynamic viewpoints. It outperforms the state-of-the-art DexNet methods by approximately 21\% for mixed cluttered scenes. In the future, we plan to study a multi-gripper grasping policy that enables swapping between different task-specific end-effectors to increase the grasp success rate and handle more challenging objects.

\section{ACKNOWLEDGEMENTS}
\vspace{-0.05in}
The authors would like to acknowledge the use of the facilities at the Indiana Next Generation Manufacturing Competitiveness Center (IN-MaC) for this paper. A portion of this work was supported by a Space Technology Research Institutes grant (\# 80NSSC19K1076) from NASA’s Space Technology Research Grants Program.  
\vspace{-0.05in}

\bibliographystyle{IEEEtran}
\bibliography{references}
\vspace{-1.5in}

\begin{IEEEbiography}
[{\includegraphics[width=1in,height=1.25in,clip,keepaspectratio]{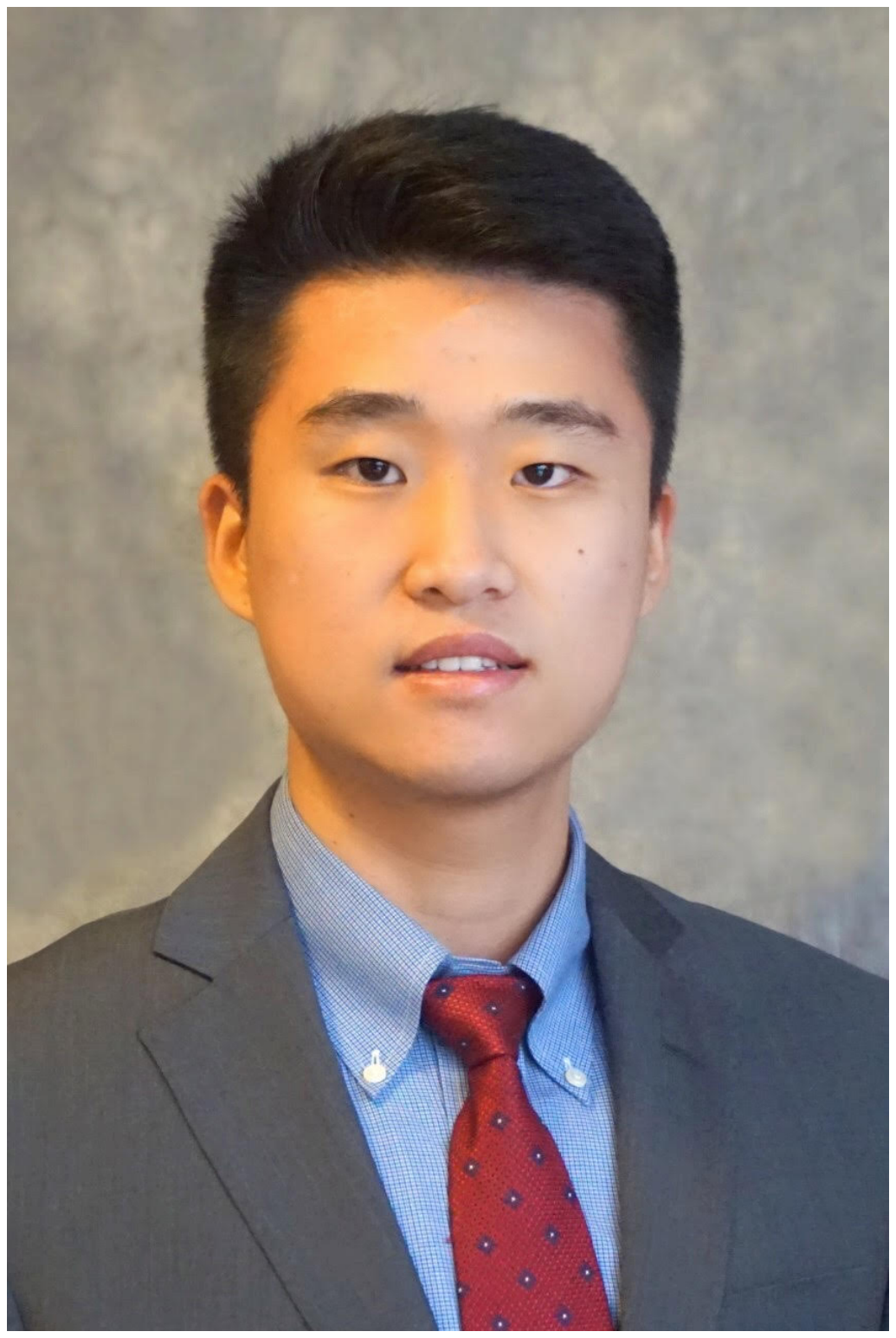}}]{Juncheng Li} received his B.E. degree in Mechanical Engineering from Stony Brook University, NY in 2018. He then obtained his M.S.E. degree in Robotics from the GRASP Lab at the University of Pennsylvania, Philadelphia, PA in 2020. Currently, he is pursuing his Ph.D. in Mechanical Engineering at the Multi-Scale Robotics and Automation Lab (MSRAL) at Purdue University, West Lafayette, IN. His research interests span grasping policy, modular end-effector systems, and computer vision.
\end{IEEEbiography}
\vspace{-1.5in}

\begin{IEEEbiography}
[{\includegraphics[width=1in,height=1.25in,clip,keepaspectratio]{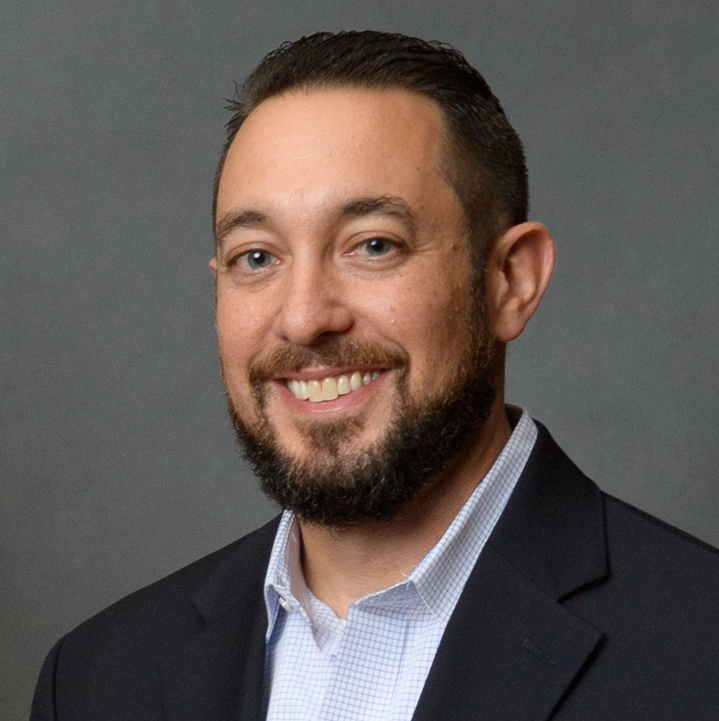}}]{David J. Cappelleri} (M’09) received a B.S. degree in mechanical engineering from Villanova University, Villanova, PA, USA,  an M.S. degree in mechanical engineering from Pennsylvania State University, State College, PA, USA, and a Ph.D. degree in mechanical engineering and applied mechanics from the University of Pennsylvania, Philadelphia, PA, USA.
He is currently a Professor with the School of Mechanical Engineering and Weldon School of Biomedical Engineering (By Courtesy) at Purdue University, West Lafayette, IN, USA and directs the Multi-Scale Robotics and Automation Lab.

Prof. Cappelleri is a recipient of the National Science Foundation CAREER Award, the Harvey N. Davis Distinguished Assistant Professor Teaching Award, the Association for Lab Automation Young Scientist Award, and the B.F.S. Schaefer Scholar Award. He is a member of the IEEE Robotics and Automation Society Technical Committee on Micro/Nano Robotics and Automation and is on the Editorial Board of IEEE Robotics and Automation Letters, and the Journal of Micro and Bio Robotics.

\end{IEEEbiography}

\end{document}